\documentclass[journal]{IEEEtran}
\usepackage{times,latexsym}
\usepackage{microtype}
\usepackage{graphicx} \graphicspath{{figures/}}
\usepackage{amsmath,amssymb,mathabx,amsbsy,mathtools,etoolbox,mathbbol}
\usepackage{algorithmic}
\usepackage[linesnumbered,ruled,vlined]{algorithm2e}
\usepackage{acronym}
\usepackage{enumitem}
\usepackage{balance}
\usepackage{xspace}
\usepackage{setspace}
\usepackage[dvipsnames,svgnames,x11names]{xcolor}
\usepackage{booktabs,tabularx,colortbl,multirow,array,makecell}
\usepackage{overpic,wrapfig}
\usepackage{nicefrac}
\usepackage[misc]{ifsym}
\usepackage{siunitx} 
\usepackage{soul} 
\usepackage[export]{adjustbox} 
\usepackage[hidelinks,colorlinks,linkcolor=blue,citecolor=blue,urlcolor=black]{hyperref}
\usepackage[accsupp]{axessibility}
\usepackage[capitalise,nameinlink]{cleveref}

\usepackage{amsmath,amsfonts}
\usepackage{algorithmic}
\usepackage{array}
\usepackage[caption=false,font=normalsize,labelfont=sf,textfont=sf]{subfig}
\usepackage{textcomp}
\usepackage{stfloats}
\usepackage{url}
\usepackage{verbatim}
\usepackage{graphicx}
\usepackage{cite}
\usepackage{color}
\usepackage{threeparttable}
\usepackage{orcidlink}

\definecolor{WeiColor}{rgb}{1,0.33,0.64}
\definecolor{myDarkBlue}{RGB}{55,81,139}
\definecolor{myLightBlue}{RGB}{158,186,217}
\definecolor{myLightGreen}{RGB}{126,171,85}
\definecolor{myRed}{RGB}{176,36,24}

\makeatletter
\DeclareRobustCommand\onedot{\futurelet\@let@token\@onedot}
\def\@onedot{\ifx\@let@token.\else.\null\fi\xspace}

\makeatother

\crefname{algorithm}{Alg.}{Algs.}
\Crefname{algocf}{Alg.}{Algs.}
\crefname{section}{Sec.}{Secs.}
\Crefname{section}{Section}{Sections}
\crefname{table}{Tab.}{Tabs.}
\Crefname{table}{Table}{Tables}
\crefname{figure}{Fig.}{Fig.}
\Crefname{figure}{Figure}{Figure}

\acrodef{cvae}[cVAE]{conditional Variational Autoencoder}
\acrodef{cgan}[cGAN]{conditional generative adversarial networks}
\acrodef{tamp}[TAMP]{task and motion planning}
\acrodef{rl}[RL]{reinforcement learning}
\acrodef{mdp}[MDP]{Markov decision process}
\acrodef{bc}[BC]{Behavior Cloning}
\acrodef{fc}[FC]{fully-connected}
\acrodef{cnn}[CNN]{convolution neural network}

\SetCommentSty{mycommfont}

\begin{document}

\title{M\textsuperscript{3}Detection: Multi-Frame Multi-Level Feature Fusion for Multi-Modal 3D Object Detection with Camera and 4D Imaging Radar}
\author{
Xiaozhi Li$^{\orcidlink{0000-0001-6699-3891}}$, Huijun Di$^{\orcidlink{0000-0002-6432-2127}}$, Jian Li$^{\orcidlink{0009-0006-0941-5861}}$, Feng Liu$^{\orcidlink{0009-0009-7351-7266}}$ and Wei Liang$^{\orcidlink{0000-0002-7539-3107}}$

\thanks{This work was supported in part by the National Natural Science Foundation of China under Grant 62172043. \textit{(Corresponding author: Huijun Di; Jian Li.)}}
\thanks{Xiaozhi Li and Jian Li are with the Radar Technology Research Institute, School of Information and Electronics, Beijing Institute of Technology, Beijing, 100081, China and with the Key Laboratory of Electronic and Information Technology in Satellite Navigation, Ministry of Education, Beijing 100081, China (e-mail: lixz@bit.edu.cn; lijian\_551@bit.edu.cn).}
\thanks{Huijun Di and Wei Liang are with the School of Computer Science, Beijing Institute of Technology, Beijing, 100081, China (e-mail: ajon@bit.edu.cn; liangwei@bit.edu.cn).}
\thanks{Feng Liu is with the Beijing Racobit Electronic Information Technology Co., Ltd., Beijing, 100081, China (e-mail: liufeng@racobit.com).}
}

\maketitle

\begin{abstract}
Recent advances in 4D imaging radar have enabled robust perception in adverse weather, while camera sensors provide dense semantic information. Fusing these complementary modalities has great potential for accurate and cost-effective 3D perception. However, most existing camera-radar fusion methods are limited to single-frame inputs, capturing only a partial view of the scene. The incomplete scene information, compounded by image degradation and 4D radar sparsity, hinders overall detection performance. In contrast, multi-frame fusion offers richer spatiotemporal information but faces two challenges: achieving robust and effective object feature fusion across frames and modalities, and mitigating the computational cost of redundant feature extraction. Consequently, we propose M\textsuperscript{3}Detection, a unified multi-frame 3D object detection framework that performs multi-level feature fusion on multi-modal data from camera and 4D imaging radar. In contrast to conventional architectures, our framework leverages intermediate features from the baseline detector and employs the tracker to produce reference trajectories, improving computational efficiency and providing richer information for second-stage. In the second stage, to address tracking uncertainties and enable fine-grained modeling, we design a global-level inter-object feature aggregation module (GOA) guided by radar information to align global features across candidate proposals and a local-level inter-grid feature aggregation module (LGA) that expands local features along the reference trajectories to enhance fine-grained object representation. The aggregated features are then processed by a trajectory-level multi-frame spatiotemporal reasoning module (MSTR) to encode cross-frame interactions and enhance temporal representation. Extensive experiments on the View-of-Delft and TJ4DRadSet datasets demonstrate that M\textsuperscript{3}Detection achieves state-of-the-art 3D detection performance, validating its effectiveness in multi-frame detection with camera-4D imaging radar fusion.
\end{abstract}

\begin{IEEEkeywords}
3D object detection, 4D imaging radar, camera, multi-frame detection, deep learning, autonomous driving.
\end{IEEEkeywords}

\section{Introduction}
\IEEEPARstart{T}{hree}-dimensional object detection is a fundamental technology in autonomous driving, aiming to identify and localize objects of interest with precise 3D bounding boxes\cite{survey_autonomous_driving}. Recent advances in 4D imaging radar have enabled long-range sensing and velocity estimation, while maintaining advantages such as all-weather robustness and cost efficiency, making it a promising sensor for 3D object detection\cite{survey_4d_radar}. However, the resolution and measurement noise of 4D imaging radar limit its ability to capture fine-grained geometric details of objects, resulting in constrained performance when used as a standalone sensor. This limitation can be effectively mitigated by fusing radar data with camera images, where the dense semantic information from the visual modality complements the spatial cues provided by radar\cite{sgdet3d,sfgfusion,hgsfusion}. 

\begin{figure}[t]
\centering{\includegraphics[scale=0.30]{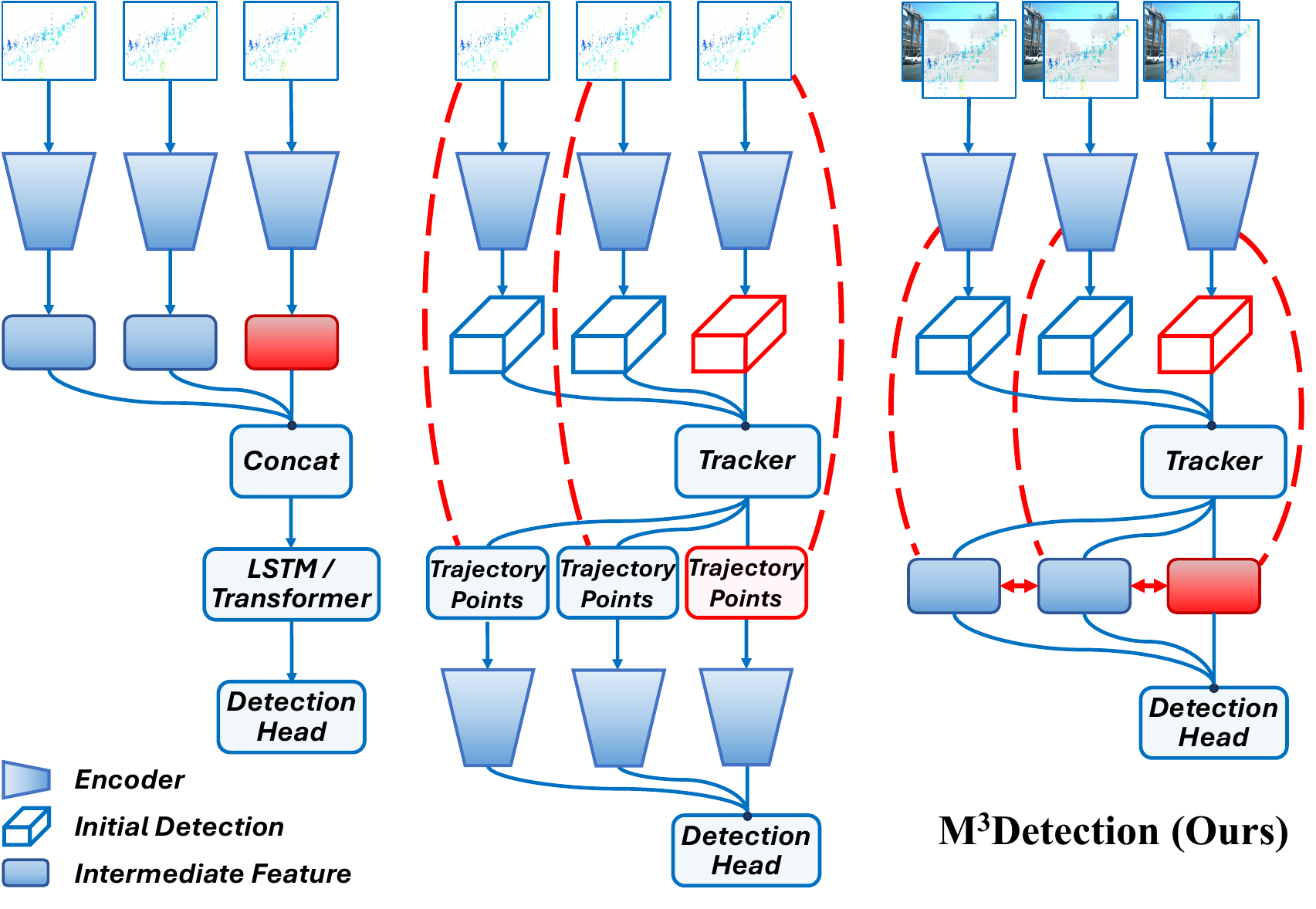}}
\caption{\textbf{Comparison of multi-frame 3D detection frameworks.} Single-stage methods apply LSTM or Transformer networks to fuse scene-level information across frames. Two-stage methods generate initial detections using a baseline detector and refine them by re-extracting features from trajectory point clouds. In contrast, our method eliminates redundant feature re-extraction and performs multi-frame multi-level feature fusion on multi-modal intermediate features, achieving enhanced performance while maintaining computational efficiency.
}
\label{Fig: M3_Methods}
\end{figure}

Despite the potential benefits offered by sensor fusion, both 4D imaging radar and camera capture only a partial view of the scene at any given time, leading to challenges in 3D object detection due to incomplete information\cite{survey_robust}. Additionally, single-frame image degradation and the sparsity of 4D radar point clouds further complicate tasks such as object location and classification. In contrast, sequential images and point clouds provide multi-view observations that embed rich spatiotemporal information. How to fully exploit the multi-view spatiotemporal cues is key to boosting 3D detection performance\cite{sequential_survey}.

In multi-frame 3D object detection, the primary challenge lies in introducing an effective multi-frame information fusion framework under limited computational budgets. The most direct approach is to merge historical point clouds with the current frame \cite{centerpoint,embracing}, which may cause trailing artifacts that degrade moving object detection. Recent methods introduce explicit temporal modeling, as illustrated in \cref{Fig: M3_Methods}, with most focusing on point cloud-only modality. Some single-stage methods extract per-frame features and directly concatenate them \cite{gmpnet-pami,centerformer,stemd}, partially capturing temporal information while retaining efficiency, but cannot fully mitigate trailing artifacts. Recent advanced two-stage methods generate per-frame proposals and refine them through trajectory point cloud feature re-extraction \cite{mppnet,once-detected,detzero}. While effective for dense LiDAR, this strategy is less applicable to 4D imaging radar because it fails to leverage global contextual information, and the local sparse trajectory point clouds lack sufficient structure for high-quality feature refinement. Meanwhile, the second-stage re-extraction introduces computational redundancy, restricting real-time performance. Furthermore, in multi-modal multi-frame detection, additional challenges arise from the differences between point clouds and images in data characteristics, which complicate feature fusion across frames. Therefore, an efficient and effective framework capable of cross-modal multi-frame feature fusion is urgently needed.

We propose an innovative two-stage multi-frame detection framework that performs spatiotemporal reasoning directly on intermediate features from the baseline detector. Specifically, we store intermediate features and initial detection results from the baseline detector, and then generate reference trajectories using a tracker to guide second-stage feature aggregation. In the second stage, corresponding features along each trajectory are aggregated across historical frames and integrated via trajectory-level spatiotemporal reasoning. This framework offers two advantages: 1) By leveraging intermediate features, the framework fully reuses the multi-modal fusion and global contextual modeling capabilities of the baseline detector, providing richer cues for second-stage refinement to mitigate the adverse impact of sparse 4D radar point clouds; 2) It eliminates the trajectory point cloud feature re-extraction step, improving computational efficiency and facilitating real-time execution.

In addition to the challenge of framework design under limited computational budgets, achieving robust object association and effective fine-grained spatiotemporal information fusion is another key challenge in multi-frame detection. In current two-stage multi-frame detection frameworks \cite{mppnet,detzero,once-detected}, object associations across frames are usually established via tracking results. Nevertheless, these methods often ignore tracking errors, such as track loss and mistracking, which can disrupt feature aggregation and compromise detection accuracy. Furthermore, existing multi-modal multi-frame detection methods \cite{fusionformer,bevfusion4d,dmfusion} generally integrate scene-level features across frames via long short-term memory (LSTM) \cite{lstm} or transformer \cite{attention}. While such strategies can partially capture temporal relations, their lack of fine-grained modeling for individual objects limits the ability to achieve precise trajectory-level temporal interaction.

To address the uncertainty in tracking and the demand for fine-grained modeling, we adopt a multi-level feature fusion strategy that separately aggregates global and local intermediate features to enhance overall feature expressiveness. Guided by 4D radar cues, a global-level inter-object feature aggregation module (GOA) combines global features from multi-hypothesis candidate proposal positions, mitigating tracking uncertainties to ensure higher feature recall and precise aggregation. In parallel, a local-level inter-grid feature aggregation module (LGA) captures richer local context through crop region expansion and cross-level deformable attention, enhancing fine-grained feature representation along the reference trajectory. After aggregation, a trajectory-level multi-frame spatiotemporal reasoning module (MSTR) employs multi-head attention to model temporal feature interactions within each trajectory, strengthening the motion awareness and temporal understanding of feature representations.

Overall, the main contributions of this work can be summarized as follows:

\begin{enumerate}
\item{We propose M\textsuperscript{3}Detection, a unified multi-frame 3D object detection framework for multi-modal fusion of camera and 4D radar data. The framework performs spatiotemporal reasoning directly on intermediate features, thereby fully reusing the modality fusion and contextual modeling capabilities of the baseline detector, and removing redundant feature re-extraction to improve computational efficiency.}
\item{We introduce a multi-frame multi-level feature fusion strategy comprising a global-level inter-object feature aggregation module to mitigate tracking uncertainties through multi-hypothesis feature alignment, a local-level inter-grid feature aggregation module to enhance fine-grained object representation through feature expansion, and a trajectory-level multi-frame spatiotemporal reasoning module to ensure integrated temporal modeling across object trajectories through multi-head attention.}
\item{Extensive experiments on the VoD and TJ4DRadSet datasets show that M\textsuperscript{3}Detection outperforms state-of-the-art methods, confirming its ability to leverage multi-level feature and temporal information for robust multi-frame 3D object detection with camera-4D imaging radar fusion.}
\end{enumerate}

The remainder of the paper is organized as follows. Section \ref{section related work} briefly reviews previous studies closely related to our work. Section \ref{Methodology} introduces the overall structure of the M\textsuperscript{3}Detection and elaborates on its key components. In Section \ref{Experiments And Analysis}, we validate the effectiveness of our method on the TJ4DRadSet and VoD datasets, followed by ablation studies to analyze the effect of different components. Finally, Section \ref{Conclusion} concludes the work and suggests future research directions.

\section{Related Work}\label{section related work}
\subsection{Single-Frame 3D Object Detection}
In autonomous driving scenarios, recent studies on 3D object detection can be categorized into three types based on the input modality: image-based \cite{imvoxelnet,BEVRefiner}, point cloud-based \cite{pointpillars,RPFA-Net,smurf}, fusion-based \cite{pv_enconet,virtual-sparse,RCBevdet}. In this study, we concentrate on the fusion-based category, which integrates semantic cues from images with geometric information from point clouds, thereby improving detection accuracy through cross-modal interactions. 

Fusion-based detectors can be further divided according to the feature integration strategies, including point feature expansion \cite{pv_enconet,pointaugmenting}, region of interest (RoI) feature fusion \cite{mv3d,f-pointnet}, virtual point feature fusion \cite{virtual-sparse,cl3d}, and BEV feature fusion \cite{bevfusion_NeurIPS,bevfusion,lss}. Early methods extend the point cloud features with image information, while others encode the two modalities separately and fuse the features at the RoI level. With advancements in depth estimation, some methods project image features into 3D space under the guidance of point clouds, generating pseudo point clouds for feature fusion. Alternatively, BEV feature fusion unifies multi-modal data into a BEV representation, enabling efficient 2D feature fusion while preserving global spatial relationships. Our proposed framework is designed for single-frame BEV feature fusion detectors and can be seamlessly adapted to most existing approaches.

\subsection{Multi-Frame 3D Object Detection}
\textbf{3D object detection from point cloud sequence data.} 
Point cloud sequences provide multiple perspectives of objects and a more comprehensive understanding of the surrounding environment. Previous studies have demonstrated that directly concatenating historical point clouds can enhance the performance of single-frame detectors \cite{centerpoint,embracing}. However, the absence of explicit temporal modeling often leads to degraded performance in long point cloud sequences, especially for fast-moving objects affected by the “tail” effect \cite{mppnet,3d-man}. Recent works employ recurrent networks to model the temporal dependencies within point cloud sequences. After extracting scene-level multi-frame features, these methods utilize LSTM, GRU, or Transformer for feature aggregation. GMPNet \cite{gmpnet-pami} introduces an attentive spatiotemporal transformer GRU to enhance object alignment through spatial and temporal attention. CenterFormer \cite{centerformer} adopts a cross-attention transformer to fuse object features across frames, while STEMD \cite{stemd} employs a spatial-temporal graph attention network to model object interactions. Although these methods effectively aggregate multi-frame information, they mainly focus on scene-level BEV features and lack explicit object-level modeling. 

To overcome this limitation, recent high-performance detectors have adopted a two-stage framework for multi-frame point cloud object detection. In the first stage, a baseline detector generates bounding boxes and velocity estimates for each frame, and a tracker associates detections across frames to construct object trajectories. The second stage refines detection results by extracting features from the original trajectory point clouds using a region-based network. MPPNet \cite{mppnet} employs proxy points in the second stage to aggregate features from the original point clouds of object trajectories. It further enhances spatiotemporal reasoning through short-clip feature fusion and whole-sequence feature aggregation. CTRL \cite{once-detected} improves trajectory tracking with a bidirectional tracking module that extends object trajectories in both directions and refines detection results using a track-centric learning module. DetZero \cite{detzero} introduces an attention-based refining module to enhance contextual feature interactions across long-term sequential point clouds, thereby improving offline detection robustness. While these approaches relatively enhance detection accuracy, they rely on trajectory point cloud feature re-extraction for second-stage refinement, which limits the use of global contextual information and introduces computational redundancy that reduces inference efficiency.

\textbf{3D object detection from multi-modal sequence data.}
Multi-modal sequential data provides a dense representation of the environment by integrating images and point clouds. However, challenges in cross-modal and cross-frame feature fusion significantly increase the complexity of object detection. Existing methods can be generally classified based on the order of cross-modal and cross-temporal fusion. Some works first process sequences independently in the image and point cloud branches to aggregate temporal features within each modality, and then project these features into a unified space for cross-modal fusion. 4D-Net \cite{4d-net} employs tiny video networks in the image branch to aggregate temporal features from multi-frame video input, while the point cloud branch utilizes a simple concatenation method to merge temporal information. The fused temporal features are then projected into a unified 3D space for cross-modal feature integration. DeepFusion \cite{deepfusion} ignores temporal dependencies in the image branch, using only the current frame, while the point cloud branch randomly discards previous frames via DropFrame. After independent feature extraction, the modalities are fused via a cross-attention module, though information loss during temporal aggregation may impede effective cross-modal alignment.

More recent approaches reverse the order by first integrating multi-modal features for each frame and then modeling temporal dependencies. Methods such as FusionFormer \cite{fusionformer}, BEVFusion4D \cite{bevfusion4d}, and DMFusion \cite{dmfusion}, LIFT \cite{lift} first extract multi-modal fused features for each frame and store them in a memory bank, after which temporal interactions are performed via 2D convolutions or attention mechanisms. Despite their effectiveness, these methods primarily focus on scene-level BEV features, overlooking refinements for individual objects, which are crucial for maintaining accuracy in cluttered environments with multiple dynamic objects. To address these limitations, our proposed M\textsuperscript{3}Detection adopts a concise two-stage framework. Instead of solely relying on scene-level temporal aggregation, M\textsuperscript{3}Detection performs multi-level feature aggregation and trajectory-level spatiotemporal reasoning on intermediate features, enabling effective multi-frame feature fusion while maintaining computational efficiency.
 
\section{Methodology}\label{Methodology}
\begin{figure*}[htbp]
\centering{\includegraphics[width=\textwidth, height=\textheight, keepaspectratio]{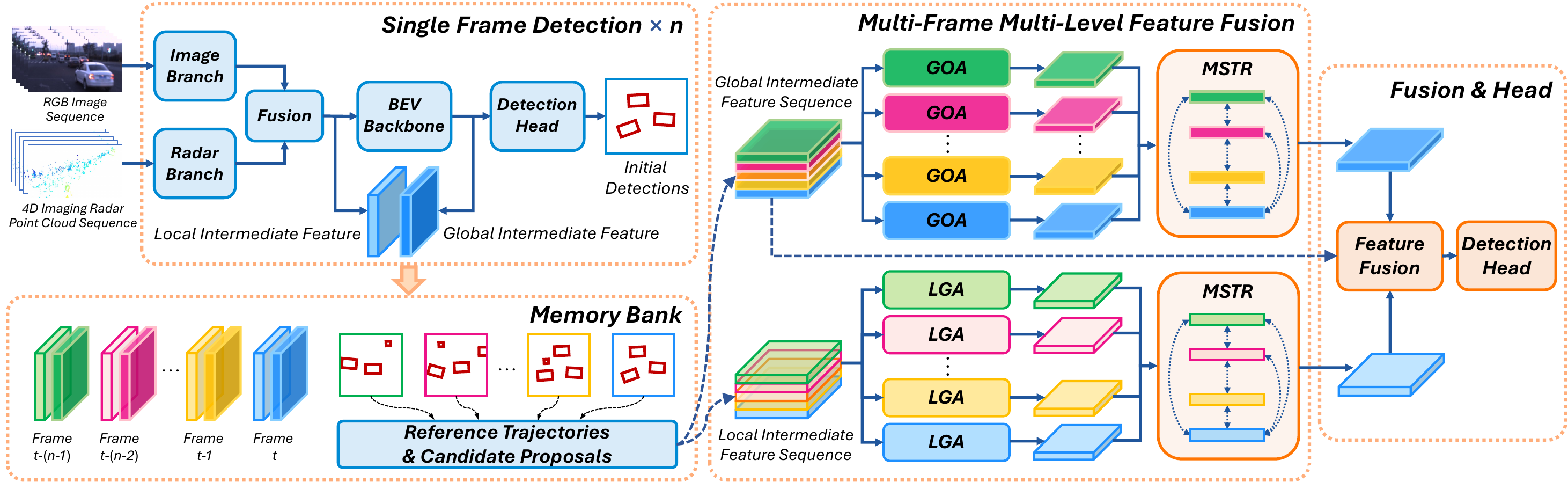}}
\caption{\textbf{The overall architecture of M\textsuperscript{3}Detection.} The framework is a two-stage pipeline for multi-frame 3D detection with camera and 4D radar. In the first stage, a single-frame multi-modal baseline detector extracts local and global intermediate features and generates initial detection results, which are associated into reference trajectories and candidate proposals by a tracking module. These features and trajectories are stored in a memory bank for the second stage, where multi-frame feature aggregation and spatiotemporal reasoning are performed. GOA aligns global features across candidate proposal positions to improve recall while preserving precision, LGA expands the crop region around reference trajectory positions and leverages cross-level deformable attention to capture richer local context, and MSTR employs multi-head attention to enable trajectory-level spatiotemporal interactions across frames. Finally, fused global and local features are used for bounding box and category regression.}
\label{Fig: Model_Architecture_ALL}
\end{figure*}

We propose M\textsuperscript{3}Detection, a unified multi-frame 3D detection framework that processes multi-modal sequences of images and 4D imaging radar point clouds through multi-level feature fusion for accurate and robust object detection. The overall architecture is illustrated in \cref{Fig: Model_Architecture_ALL}.

In the first stage, the images and 4D radar point clouds are fed into a single-frame multi-modal 3D baseline detector to extract local and global intermediate features in the BEV space and generate initial detection results. These results are then accessed by a tracking module to generate both reference trajectories and candidate proposals. The resulting tracking information and intermediate features are stored in the memory bank, which subsequently serves as input to the second stage for further refinement.

In the second stage, multi-frame and multi-level feature fusion is performed based on the input. As illustrated in \cref{Fig_Single_Multi_Tracking}, the global-level inter-object feature aggregation module (GOA) is applied to global intermediate features by aligning and integrating features from multi-hypothesis candidate proposal positions, mitigating tracking uncertainties to ensure higher feature recall and precision. In parallel, the local-level inter-grid feature aggregation module (LGA) is applied to local intermediate features based on reference trajectory positions, using crop region expansion and cross-level deformable attention to enhance fine-grained feature representation and enlarge the receptive field at low-level feature layers. After aggregation, the trajectory-level multi-frame spatiotemporal reasoning module (MSTR) leverages multi-head attention to model temporal feature interactions along each object trajectory, thereby enhancing motion representation and temporal understanding. Finally, the fusion and head module fuses global and local temporal features for accurate bounding box and category prediction.

\subsection{Single-Frame Baseline Detector and Tracking}

\subsubsection{Camera-4D Imaging Radar Fusion Baseline Detector}
Our single-frame baseline detector leverages the BEVFusion framework \cite{bevfusion} to facilitate effective interaction across multi-modal inputs. The image and radar branches first extract modality-specific features, which are then projected into a unified BEV space to produce local intermediate features $F_{BEV}^{L}$ that retain both semantic and geometric information. These features are further extracted by a BEV backbone, which aggregates multi-scale contextual cues to derive global intermediate features $F_{BEV}^{G}$. The detection head subsequently generates 3D bounding boxes, serving as initial detections $P$ for downstream processing. Notably, our multi-frame detection framework is inherently model-agnostic and thus applicable to a wide range of BEV-based single-frame fusion detectors without requiring architectural modifications.

\begin{figure}[t]
\centering{\includegraphics[scale=0.3]{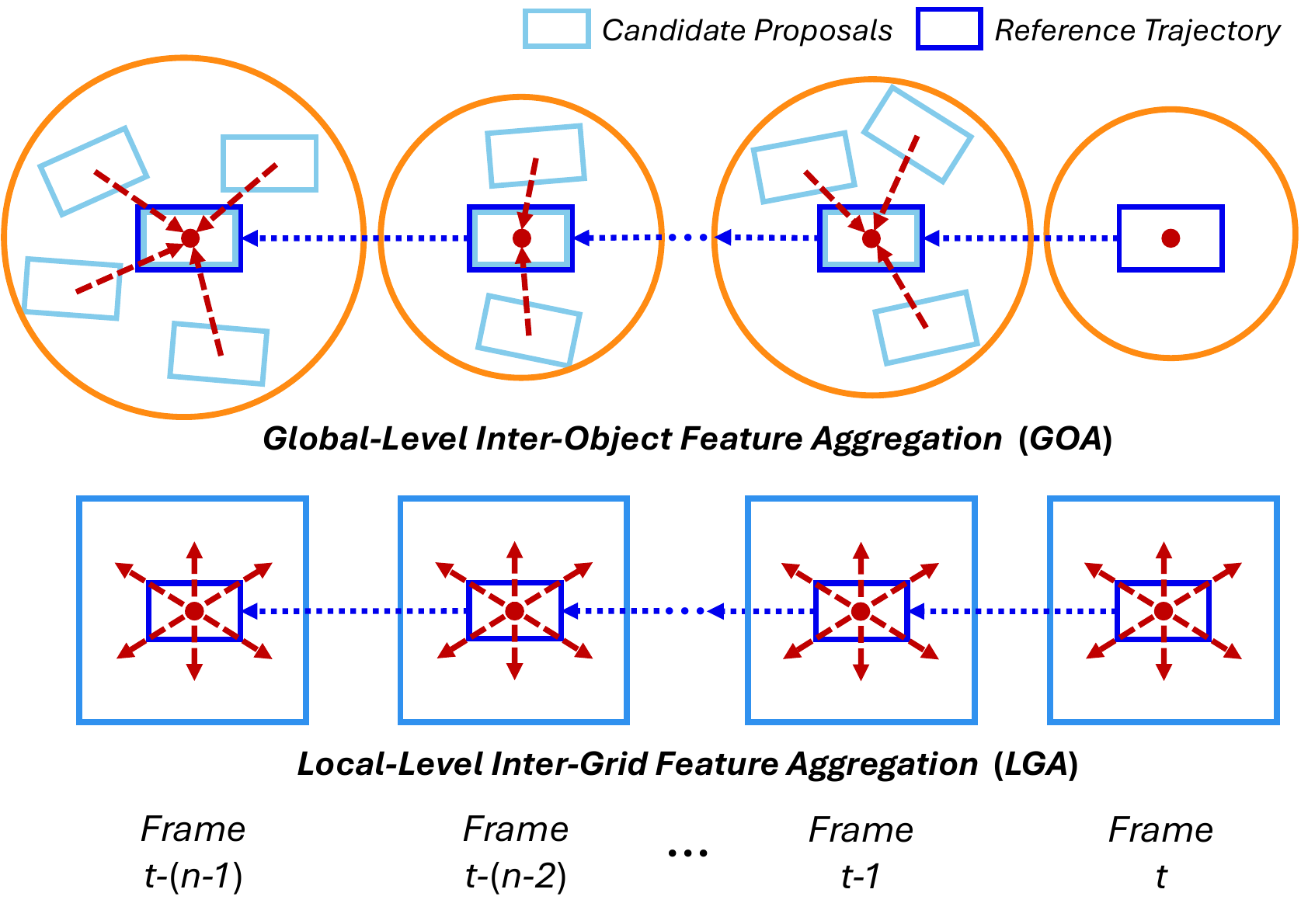}}
\caption{\textbf{Multi-level feature aggregation based on reference trajectory and candidate proposals.} GOA aligns and integrates features at candidate proposal positions to increase recall while maintaining feature precision. LGA performs crop region expansion and cross-level deformable attention along the reference trajectory to enrich local context and enlarge the receptive field.}
\label{Fig_Single_Multi_Tracking}
\end{figure}

\begin{figure}[t]
\centering{\includegraphics[scale=0.44]{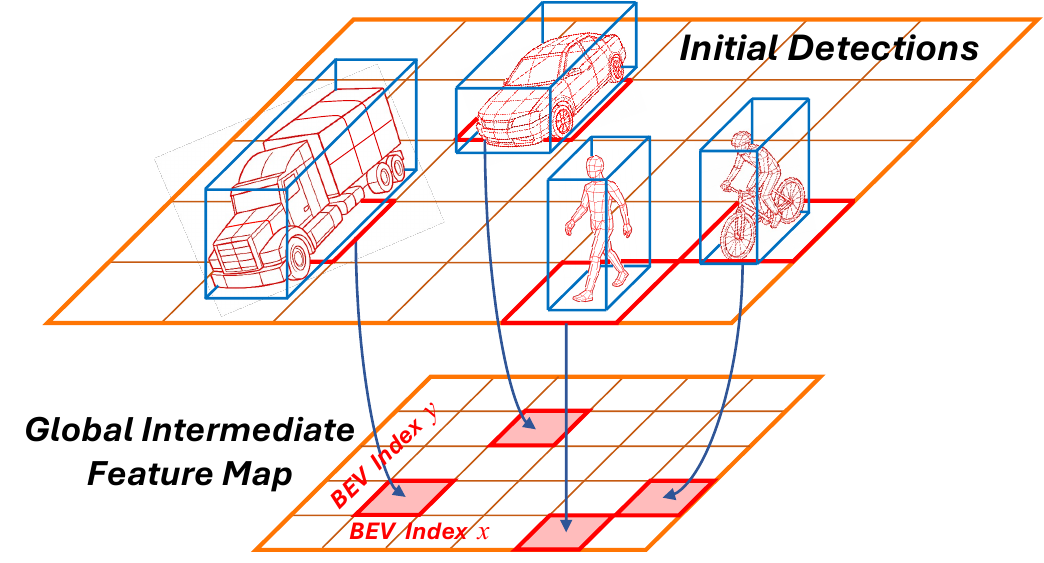}}
\caption{\textbf{Spatial correspondence between the initial detections and BEV indexes.} Each 3D bounding box is spatially aligned with the global BEV representation during single-frame inference, enabling its associated feature values to be directly retrieved using the recorded BEV indexes.}
\label{Fig_Detection-BEV-index_Result}
\end{figure}

To enhance the overall efficiency of the algorithm, we eliminate the redundant feature re-extraction typically involved in traditional multi-frame detection methods. Instead, we utilize the tracking results to directly crop the corresponding historical intermediate features for each trajectory, based on the spatial relationship between the initial detections and the intermediate features in each frame. Accordingly, during the single-frame detection, we store not only the detection results but also the BEV indexes $I_{BEV}$ of the corresponding bounding boxes in the global intermediate feature map along the $x$ and $y$ dimensions, as shown in \cref{Fig_Detection-BEV-index_Result}. This facilitates efficient alignment between tracking trajectories and historical intermediate features in the subsequent multi-frame feature aggregation.

\subsubsection{Memory Bank}
We introduce a memory bank to store historical intermediate features and initial detections for subsequent object tracking and spatiotemporal feature fusion. The memory bank $S$, holds $n$ frames of data, with each frame containing global intermediate features $F_{BEV}^{G}$, local intermediate features $F_{BEV}^{L}$, initial detections $P$, and BEV indexes $I_{BEV}$. The memory bank follows a first-in, first-out (FIFO) update strategy. 

Specifically, at time $t$, we perform tracking using the initial detections stored from the past $n$ frames in the memory bank. Guided by the tracking results, historical intermediate features are aggregated across frames to refine the detection for the time $t$. Afterward, we pop the oldest historical data $S_{t-(n-1)}$ from the memory bank and, at the beginning of the next detection cycle, insert the newly extracted data $S_{t+1}$ corresponding to the time $t+1$.

\subsubsection{Reference Trajectory and Candidate Proposals}
To achieve efficient multi-frame information integration, we perform feature aggregation and spatiotemporal reasoning at the trajectory-level. This requires associating the same object across different frames using a tracking algorithm. In this work, the Immortal tracking algorithm \cite{immortal} is adopted and extended, with enhancements specifically introduced in the detection-prediction association module. By refining the association strategy between current detections and predicted states from the previous frame, the algorithm is capable of generating both reference trajectory and candidate proposals, as illustrated in \cref{Fig_Single_Multi_Tracking}. The detailed tracking procedure is described as follows:

For each frame, we choose a relatively permissive score threshold during the single-frame detection post-processing to ensure higher recall of detection results. Subsequently, a 3D Kalman Filter (3DKF) is is utilized for trajectory prediction, where the the state vector is defined in world coordinates as $Z=[x,y,z,\theta,l,w,d,\dot{x},\dot{y},\dot{z}]$ representing the object’s position, size, orientation, and velocity. For detection-prediction association, we compute the 3D GIoU between detected bounding boxes and boxes predicted by the 3DKF, and perform Hungarian matching accordingly. During matching, any pair with a 3D GIoU below the threshold $\text{GIoU}_\text{thr\_low}$ is discarded, and the unique assignment produced by Hungarian matching is taken as the single-hypothesis reference trajectory. In addition, detection-prediction pairs with GIoU exceeding a higher threshold $\text{GIoU}_\text{thr\_high}$ are selected for multi-hypothesis candidate proposals; if more than five matches are found, the top five with the highest detection scores are retained. The Kalman filter state is then updated based on the reference trajectory matches, while unmatched detections are used to initialize new tracks. 

\subsection{Multi-Frame Multi-Level Feature Fusion}
Multi-frame sequence data inherently contains object features captured from multiple views, offering enriched information for more accurate 3D bounding box estimation. However, aggregating effective features for each object across long sequences remains challenging due to the large number of proposals and their dynamic spatial distributions. To address the challenge, we perform multi-frame feature extraction at the trajectory level,  aggregating features and conducting temporal reasoning for each trajectory independently. The pipeline comprises two stages: single-frame feature aggregation and multi-frame spatiotemporal reasoning. In the first stage, the inter-object feature aggregation module and inter-grid feature aggregation module are applied to global and local intermediate features, respectively, producing a compact representation for each object within a single frame. Subsequently, these aggregated features serve as input to the trajectory-level multi-frame spatiotemporal reasoning module, which enables spatiotemporal interaction along each trajectory, yielding enriched multi-frame features for accurate object geometry and category regression.

\begin{figure}[t]
\centering{\includegraphics[scale=0.378]{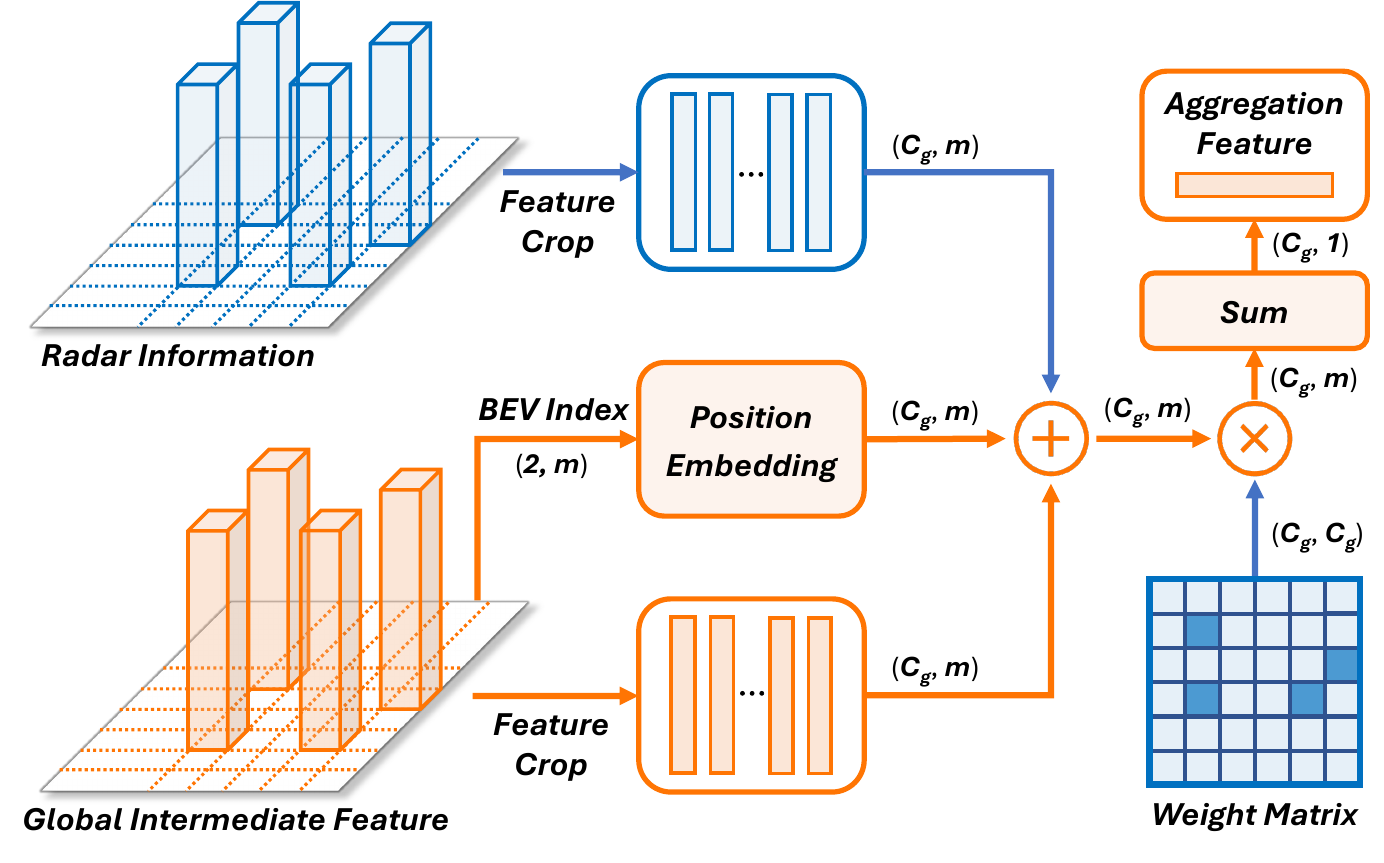}}
\caption{\textbf{Global-level inter-object feature aggregation (GOA).} Based on candidate proposals, BEV features are aggregated from candidate positions to enrich the representation. Feature reliability is ensured by leveraging radar-aware features, spatial distribution of candidate proposals, and an adaptive weight matrix for robust aggregation.
}
\label{Fig_Globla_Feature_Aggregation}
\end{figure}

\subsubsection{Global-Level Inter-Object Feature Aggregation}
As the final input to the detection head, the global intermediate feature encodes highly abstract object information while maintaining strict spatial alignment with the geometry of the real-world scene. As illustrated in \cref{Fig_Globla_Feature_Aggregation}, the global-level inter-object feature aggregation module (GOA) is designed to mitigate tracking uncertainties and consolidate as much valid global-level BEV information as possible for each object within a single frame. Specifically, the module aims to maximize the recall of relevant information while simultaneously enhancing the precision of aggregated features. However, this process inherently involves a trade-off. To reduce the risk of missing object information caused by track loss or mistracking, we aggregate features from multi-hypothesis candidate proposals and relax the confidence threshold in the single-frame detection stage to preserve potential object cues in low-score regions. While these strategies effectively improve the chance of capturing useful information, they also introduce redundancy and noise, which may compromise the accuracy and discriminability of the aggregated global features.

To address this challenge, we introduce several mechanisms to preserve high-quality information selectively. First, we leverage the spatial response and multi-dimensional attributes (velocity and RCS) of 4D imaging radar to guide the selection of candidate proposals and suppress irrelevant regions. Second, we incorporate position embedding to encode the spatial layout of candidate proposals, thereby enhancing the geometric representation of aggregated features. Finally, we introduce a learnable feature selection matrix that adaptively weights each candidate feature, enabling effective fusion of the most informative representations.

Specifically, to effectively leverage 4D imaging radar information, we employ a lightweight pillar-based encoder to extract radar-aware features $F_r^g$, aligned in resolution with the global intermediate feature. The embedded velocity channel enhances sensitivity to dynamic objects and suppresses background clutter, while the RCS channel improves the discrimination of objects with complex structures or diverse materials. These enriched radar features provide effective priors for robust inter-object feature aggregation.

Given the global intermediate feature $F_{BEV}^{G} \in \mathbb{R}^{C_g \times H_g \times W_g}$ and radar-aware feature $F_r^g \in \mathbb{R}^{C_g \times H_g \times W_g}$. We utilize the BEV indexes $I_{BEV}^{c}$ of candidate proposals to determine the spatial alignment with feature maps, and crop the features corresponding to the proposals as follows:
\begin{equation}
\label{global BEV feature crop}
f_{BEV}^{G} =\text{Crop}(F_{BEV}^{G}, I_{BEV}^{c}),
\end{equation}
\begin{equation}
\label{radar feature crop}
f_{r}^{g} =\text{Crop}(F_{r}^{g}, I_{BEV}^{c}),
\end{equation}
where $f_{BEV}^{G}\in \mathbb{R}^{C_g \times m}$ and $f_{r}^{g}\in \mathbb{R}^{C_g \times m}$ denote the cropped global intermediate features and radar-aware features corresponding to the candidate proposals, $C_g$ is the number of global feature channels and $m$ is the number of candidate proposals, $\text{Crop}(\cdot)$ represents the spatial sampling operation.

To incorporate the spatial distribution information of candidate proposals and further enhance the spatial expressiveness of the aggregated features, we apply position embedding to the BEV indexes associated with the proposals. This process can be formulated as:
\begin{equation}
\label{position embedding}
p_e =\text{PE}(I_{BEV}^{c}),
\end{equation}
where $p_e \in \mathbb{R}^{C_g \times m}$ represents the spatial positional features derived from the BEV indices, and $\text{PE}(\cdot)$ denotes the position embedding function.

Following feature cropping and position embedding, we perform element-wise addition of the cropped global intermediate features $f_{BEV}^{G}$, radar-aware features $f_{r}^{g}$, and spatial positional features $p_e$, resulting in enhanced multi-modal representations. To further refine the feature quality, we introduce a learnable linear transformation matrix $W_a \in \mathbb{R}^{C_g \times C_g}$, which projects and reconstructs each fused feature vector:
\begin{equation}
\label{transformation weight matrix}
f^a =(f_{BEV}^{G} + f_{r}^{g} + p_e) \times W_a,
\end{equation}
where $f^a\in \mathbb{R}^{C_g \times m}$ is the reconstructed features of candidate proposals. The learnable matrix $W_a$ is optimized end-to-end during training, allowing the model to adaptively emphasize informative features while suppressing redundant noise. Finally, we aggregate all $m$ features via summation to produce the final per-frame object representation:
\begin{equation}
\label{featur sum}
F_A^G =\sum_{i=1}^{m}f^a_i,
\end{equation}
where $F_A^G\in \mathbb{R}^{C_g \times 1}$ is the per-frame global-level aggregation feature. By performing this aggregation for each historical frame, the complete global-level inter-object features corresponding to the individual object trajectory are obtained, integrating both semantic and spatial information from candidate proposals and providing a robust representation for subsequent spatiotemporal modeling.

\begin{figure}[t]
\centering{\includegraphics[scale=0.34]{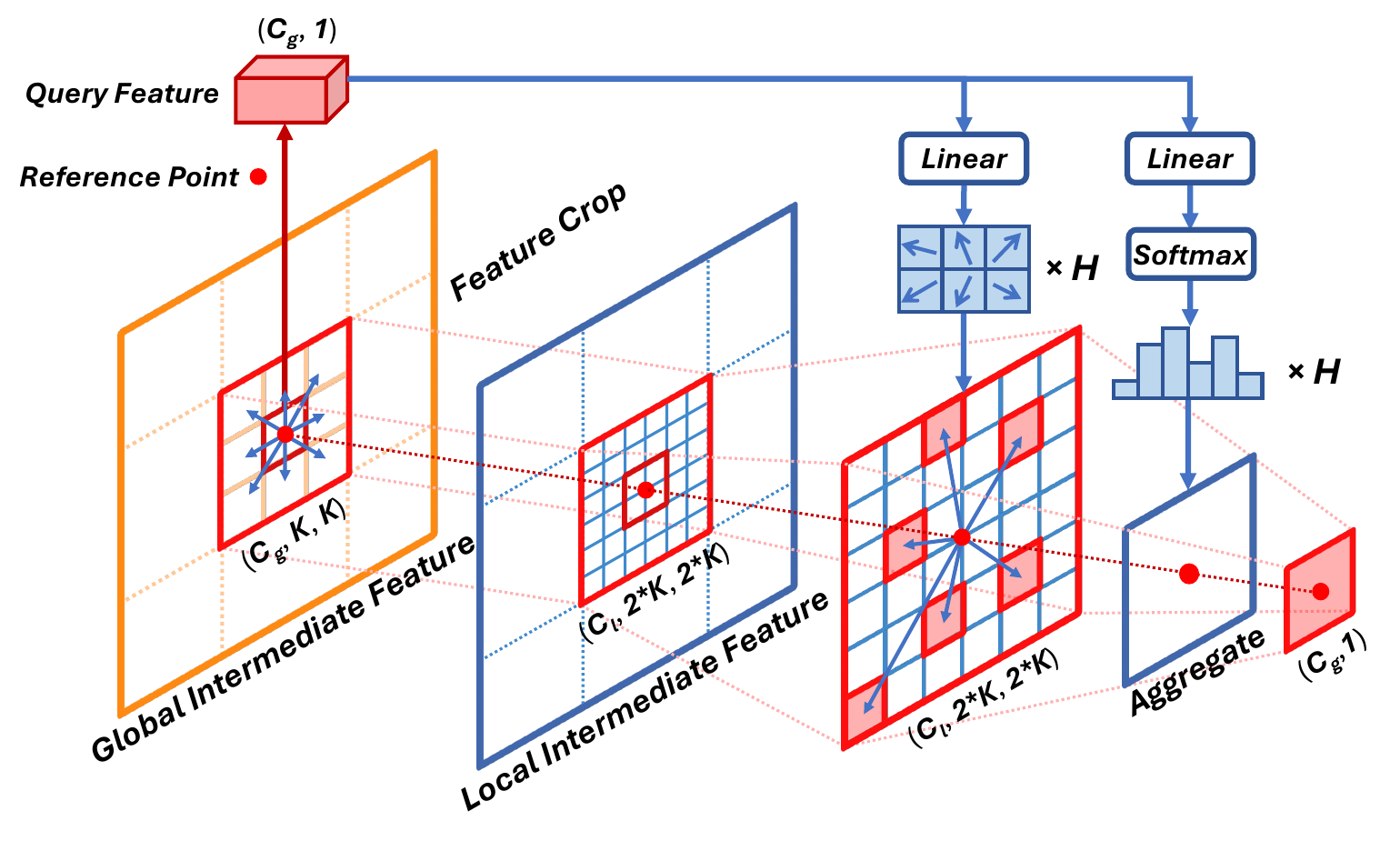}}
\caption{\textbf{Local-level inter-grid feature aggregation (LGA).} For each reference trajectory position, local intermediate features are cropped within an expanded region to capture the surrounding context. Cross-level deformable attention is applied, using the corresponding global feature as the query to guide multi-head sampling and weighting.}
\label{Fig_Local_Feature_Aggregation}
\end{figure}

\subsubsection{Local-Level Inter-Grid Feature Aggregation}
Compared to the global intermediate feature, the local intermediate feature retains higher spatial resolution and preserves more structural details, which are critical for fine-grained object detection. 

In the single-frame baseline detector, the local intermediate feature maintains twice the spatial resolution of the global intermediate feature, while both features cover the same physical area. To further capture fine-grained contextual cues, the local-level inter-grid feature aggregation module (LGA) expand the crop region around each reference trajectory position and introduce the cross-level deformable attention mechanism for efficient feature extraction, as illustrated in \cref{Fig_Local_Feature_Aggregation}. The cropped feature from the expanded region can be formally expressed as:
\begin{equation}
\label{local BEV feature crop}
f_{BEV}^{L} =\text{Crop}(F_{BEV}^{L}, I_{BEV}^{r}),
\end{equation}
where $f_{BEV}^{L} \in \mathbb{R}^{C_l \times 2K \times 2K}$ denotes the cropped local intermediate features corresponding to the expanded crop region, $C_l$ is the number of local feature channels, $K$ is the expansion factor that controls the spatial extent, $I_{BEV}^{r}$ is the BEV indexes of reference trajectory.

After feature cropping, we introduce a cross-level deformable attention mechanism to further enhance the cropped local features $f_{BEV}^{L}$. Unlike conventional deformable attention, our design leverages the global feature corresponding to the reference point $p_q$ as the query $f_q^G$. Guided by the global semantic cues, the linear layers infer sampling offsets and attention weights, facilitating cross-level interaction and enhancing feature representation. To strengthen modeling capacity, the module employs $H$ attention heads, where each head attends to distinct spatial and semantic subspaces. During feature aggregation, outputs from multiple attention heads are combined, and the channel dimension is projected from $C_l$ to $C_g$ to ensure consistency with subsequent modules. The cross-level deformable attention can be formulated as:
\begin{equation}
\begin{split}
\text{CL-DeformAttn}(&f_q^G,p_q,f_{BEV}^L) 
= \\\sum_{h=1}^{H} W_h \Bigg[ 
&\sum_{j=1}^{J} A_{hqj} \cdot W'_h \, f_{BEV}^L\big(p_q + \Delta p_{hqj}\big) 
\Bigg]
\end{split},
\end{equation}
where $h$ indexes the attention heads, $H$ is the total attention head number, $j$ indexes the sampling points, $J$ is the total sampling point number, $A_{hqj}$ and $\Delta p_{hqj}$ denote the attention weight and sampling offset of the $j^{th}$ sampling point in the $h^{th}$ attention head, respectively. The final per-frame object representation can be obtained as:
\begin{equation}
\label{crop feature extraction}
F_{A}^{L} =\text{Linear}(\text{CL-DeformAttn}(f_q^G,p_q,f_{BEV}^L)),
\end{equation}
where $F_A^L \in \mathbb{R}^{C_g \times 1}$ is the per-frame local-level aggregation feature. By applying the feature aggregation to each historical frame, the complete local-level inter-grid features along the individual object trajectory are obtained. These features emphasize fine-grained local cues and provide a refined representation for subsequent temporal modeling of object details.

\subsubsection{Trajectory-Level Multi-Frame Spatiotemporal Reasoning}
\begin{figure}[t]
\centering{\includegraphics[scale=0.53]{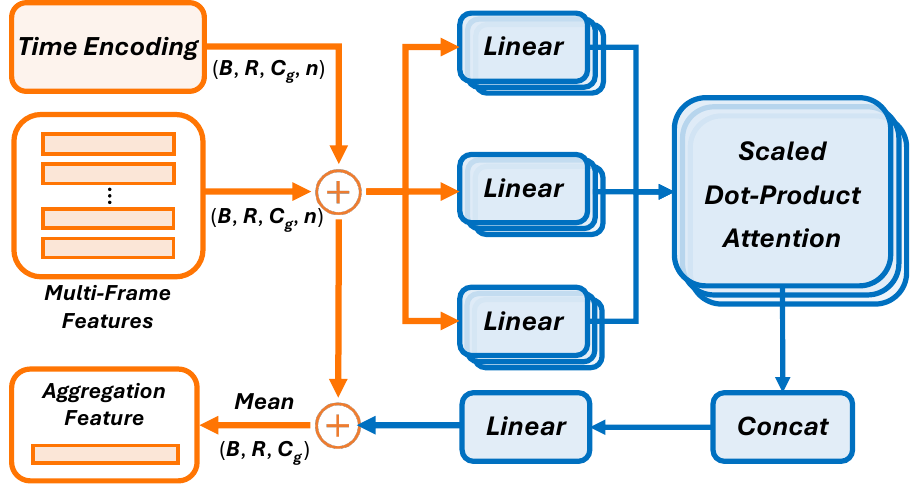}}
\caption{\textbf{Trajectory-level multi-frame spatiotemporal reasoning (MSTR).} The module employs multi-head attention with time encoding to integrate trajectory-level features across frames, yielding robust representations that enhance the stability and reliability of subsequent classification and localization.}
\label{Fig_Multi_Frame_Spatiotemporal_Reasoning}
\end{figure}

After completing the global-level and local-level feature aggregation within a single frame, we further incorporate multi-frame temporal information to enhance the completeness and robustness of object representations. Single-frame features are inherently limited to capturing momentary snapshots of object states, neglecting the temporal continuity and inter-frame context critical for reliable detection. To address this, we introduce the trajectory-level multi-frame spatiotemporal reasoning module (MSTR) as illustrated in \cref{Fig_Multi_Frame_Spatiotemporal_Reasoning}, which aggregates features within the object trajectory and leverages historical context for enhanced current-frame detection.

In the spatiotemporal reasoning stage, we employ multi-head attention (MHA) to separately model the global-level and local-level multi-frame features of each trajectory, enabling efficient spatiotemporal interaction. By focusing on trajectory-level representations, each attention unit operates on the same object across frames, which preserves object-specific context to prevent cross-object interference and enhances the motion and spatial representations of each object.

For clarity, we detail the implementation using the global-level feature branch as an example. The aggregated features of $R$ object trajectories across $n$ frames are concatenated along the temporal axis, producing a sequence $F_s \in \mathbb{R}^{B \times R \times C_g \times n}$. This tensor is then augmented with time encoding and reshaped into $F_s^{\prime} \in \mathbb{R}^{(B \cdot R) \times C_g \times n}$ to match the input format required by the attention module, which subsequently performs inter-frame feature interaction:
\begin{equation}
\label{feature MHA}
F_{attn} = \text{MHA}(\text{Reshape}(F_s+p_t)),
\end{equation}
where $\text{MHA}$ denotes the multi-head attention module, $p_t\in \mathbb{R}^{B \times R \times C_g \times n}$ is the time encoding, $F_{\text{attn}} \in \mathbb{R}^{(B \cdot R) \times C_g \times n}$ is the output feature, $B$ is the batch size and $R$ is the number of object trajectories. We add the attention output back to the input using a residual connection, then average the result across the time dimension. After removing redundant axes and reshaping, we obtain the final multi-frame aggregated feature:
\begin{equation}
\label{multi-frames feature aggregation}
F_S^G = \text{Reshape}(\text{Mean}(F_s' + F_{attn})),
\end{equation}
where $F_S^G \in \mathbb{R}^{B \times R \times C_g}$ denotes the global-level multi-frame aggregated feature. This representation provides robust support for the following object classification and localization. Similarly, the local-level feature branch undergoes the same reasoning process to produce the aggregated feature $F_S^L \in \mathbb{R}^{B \times R \times C_g}$.
 
\subsection{Fusion and Head}
\subsubsection{Feature Fusion}
Following the extraction of the multi-frame global feature $F_S^G$ and local feature $F_S^L$, we utilize the BEV indexes $I_{\text{BEV},t}$ at the current frame $t$ to reproject the features into the corresponding positions in BEV space. This process yields a unified spatial representation for downstream regression. The operation is formally defined as:
\begin{equation}
F_{BEV,s}^G = \text{Reproject}(F_S^G,I_{BEV,t}),
\end{equation}
\begin{equation}
F_{BEV,s}^L = \text{Reproject}(F_S^L,I_{BEV,t}),
\end{equation}
where $F_{BEV,s}^G$ and $F_{BEV,s}^L$ denote the BEV representations of the global and local aggregated features, respectively. These representations share the same spatial resolution and alignment as the original global intermediate feature $F_{BEV,t}^G \in \mathbb{R}^{C_g \times H_g \times W_g}$ at the current frame $t$, ensuring spatial consistency through the backward mapping process. Furthermore, we concatenate the two BEV representations with the global intermediate feature $F_{BEV,t}^G$ along the channel dimension. Here, $F_{BEV,t}^G$ provides the current-frame BEV information, which is enhanced through integration with temporal features to improve the representation of objects in the current frame. The concatenated feature is further processed by a CBR module, which can be defined as:
\begin{equation}
F_{BEV} = \text{CBR}(\text{Concat}(F_{BEV,s}^L,F_{BEV,s}^G,F_{BEV,t}^G)),
\end{equation}
where CBR denotes the sequential application of Convolution, Batch Normalization, and ReLU activation,  $F_{BEV}$ denotes the final aggregated BEV representation that incorporates multi-frame and multi-level information. This feature achieves unified modeling across the spatiotemporal domain, retaining global semantic context and fine-grained local details to support accurate object detection.

\subsubsection{Detection Head and loss}
The detection head follows an anchor-based design \cite{pointpillars}, producing 3D bounding box predictions from the aggregated BEV feature. The overall optimization objective is formulated as a weighted sum of three components: localization loss $L_{loc}$, classification loss $L_{cls}$, and orientation loss $L_{dir}$. The final training loss is expressed as:
\begin{equation}
L = \beta_1L_{loc} + \beta_2L_{cls} + \beta_3L_{dir},
\end{equation}
where $\beta_1$, $\beta_2$, and $\beta_3$ are balancing coefficients that control the relative contribution of each loss term.

\section{Experiments And Anslysis}\label{Experiments And Analysis}
\subsection{Dataset Setup and Metric}
\textbf{Dataset:} We evaluate our model on two representative and challenging 4D radar benchmarks: VoD \cite{vod} and TJ4DRadSet \cite{tj4dradset}. Both datasets are specifically designed for autonomous driving and provide synchronized multi-modal data, including monocular camera images and 4D radar point clouds. Additionally, intrinsic parameters and sensor-to-sensor transformation matrices are provided to facilitate multi-modal fusion. Unlike datasets that consist of independent samples, both TJ4DRadSet and VoD contain continuous frame sequences, which not only enable single-frame 3D object detection but also support the study of multi-frame detection strategies such as M\textsuperscript{3}Detection.

The VoD dataset was collected in Delft, Netherlands, and primarily covers campus, suburban, and old-town areas, with particular emphasis on scenarios involving vulnerable road users such as pedestrians and cyclists. It contains a total of 8,682 frames with over 123,000 annotated 3D bounding boxes across three object categories: car, pedestrian, and cyclist. The official split comprises 5,139 frames for training, 1,296 frames for validation, and 2,247 frames for testing. For evaluation, we utilize the official testing code provided by the dataset to generate the final results.

The TJ4DRadSet dataset was collected in Suzhou, China, across various road types, including urban roads, elevated highways, and industrial zones, while also presenting challenging conditions such as nighttime, glare, and camera defocus. It comprises a total of 7,757 frames across 44 sequences of synchronized camera and 4D radar data, with 5,717 frames designated for training and 2,040 for testing. The dataset provides 3D annotations for four object categories: car, pedestrian, cyclist, and truck, offering a comprehensive benchmark for 3D object detection in complex driving scenarios.

\textbf{Evaluation metrics:}
The performance of M\textsuperscript{3}Detection is assessed using the specified evaluation settings of each dataset. In the VoD dataset, evaluation follows the official protocol, which reports Average Precision (AP) over the entire annotated area and a predefined region of interest (RoI). The RoI corresponds to the ego-vehicle’s driving corridor in the camera coordinate system, bounded by $D_{RoI} = \{ (x,y,z) \mid -4m < x < 4m, z < 25m \}$. The Intersection over Union (IoU) thresholds used for AP calculation are set to 0.5 for cars, 0.25 for pedestrians, and 0.25 for cyclists.

For the TJ4DRadSet dataset, evaluation is performed within a 70-meter range from the radar sensor. The metrics include both 3D AP and BEV AP, reported for each object category as well as the mean Average Precision (mAP) across categories. The IoU thresholds follow the configuration as in the VoD dataset, with values of 0.5 for cars and trucks, and 0.25 for pedestrians and cyclists.

\definecolor{lightblue}{RGB}{220,230,245}
\begin{table*}[t]
    \centering
    \caption{Comparison of detectors' results on the VoD dataset\cite{vod}, where R denotes 4D imaging radar and C represents camera. Bold and underlined values denote the best and second-best performance, respectively}
    \label{Tab:VoD Result}
    \begin{threeparttable}
    \fontsize{10pt}{13pt}\selectfont
    \setlength{\tabcolsep}{3.8pt}
    \begin{tabular}{c|c|c|>{\centering\arraybackslash}p{1.2cm} >{\centering\arraybackslash}p{1.2cm} >{\centering\arraybackslash}p{1.2cm}|>{\centering\arraybackslash}p{1.2cm}|>{\centering\arraybackslash}p{1.1cm} >{\centering\arraybackslash}p{1.1cm} >{\centering\arraybackslash}p{1.1cm}|>{\centering\arraybackslash}p{1.1cm}|>{\centering\arraybackslash}p{0.8cm}}
    \bottomrule[1.5pt]
    \multirow{2}{*}{\centering Method} & \multirow{2}{*}{\centering Modality} & \multirow{2}{*}{\centering Year} & \multicolumn{4}{c|}{AP in the Entire Annotated Area (\%)} & \multicolumn{4}{c|}{AP in the Region of Interest (\%)} & \multirow{2}{*}{\centering FPS} \\
    \cline{4-11}
         & & & Car & Ped & Cyc & mAP & Car & Ped & Cyc & mAP & \\
    \hline
        ImVoxelNet\cite{imvoxelnet} & C & 2022 & 19.35 & 5.62 & 17.53 & 14.17 & 49.52 &  9.68 & 28.97 & 29.39 & 11.1 \\
    \hline
        PointPillars\cite{pointpillars} & R & 2019 & 38.69 & 31.54 & 65.52 & 45.25 & 71.47 & 41.54 & 87.88 & 66.97 & 42.9 \\
         PillarNeXt\cite{pillarnext} & R & 2021 &30.81 & 33.11 & 62.78 & 42.23 & 66.72 & 39.03 & 85.08 & 63.61 & 28.0 \\
         CenterPoint\cite{centerpoint} & R & 2021 & 32.74 & 38.00 & 65.51 & 45.42 & 62.01 & 48.18 & 84.98 & 65.06 & 34.5 \\
         RadarPillarNet\cite{RCFusion} & R & 2023 & 39.30 & 35.10 & 63.63 & 46.01 & 71.65 & 42.80 & 83.14 & 65.86 & \textit{N/A} \\
         SMIFormer\cite{smiformer} & R & 2023 &39.53 & 41.88 & 64.91 & 48.77 & 77.04 & 53.40 & 82.95 & 71.13 & 16.4 \\
         SMURF\cite{smurf} & R & 2024 & 42.31 & 39.09 & 71.50 & 50.97 & 71.74 & 50.54 & 86.87 & 69.72 & 23.1 \\
         SCKD\cite{sckd} & R & 2025 &41.89 & 43.51 & 70.83 & 52.08 & 77.54 & 51.06 & 86.89 & 71.80 & 39.3 \\
    \hline
         BEVFusion\cite{bevfusion} & C+R & 2023 &42.02 & 38.98 & 67.54 & 49.51 & 72.23 & 48.67 & 85.57 & 69.02 & 8.4\\
         RCFusion\cite{RCFusion} & C+R & 2023 &41.70 & 38.95 & 68.31 & 49.65 & 71.87 & 47.50 & 88.33 & 69.23 & 4.7\\
         FUTR3D\cite{futr3d} & C+R & 2023 & 46.01 & 35.11 & 65.98 & 46.84 & 78.66 & 43.10 & 86.19 & 69.32 & 7.3\\
         RCBEVDet\cite{RCBevdet} & C+R & 2024 & 40.63 & 38.86 & 70.48 & 49.99 & 72.48 & 49.89 & 87.01 & 69.80 & \textit{N/A}\\  
         LXL\cite{LXL} & C+R & 2024 & 42.33 &  49.48 &  77.12 & 56.31 & 72.18 & 58.30 & 88.31 & 72.93 & 6.1\\
         SGDet3D\cite{sgdet3d} & C+R & 2024 & 53.16 & 49.98 & 76.11 & 59.75 & 81.13 & 60.91 & \textbf{90.22} & 77.42 & 9.2 \\
         ZFusion\cite{zfusion} & C+R & 2025 & 43.89 & 39.48 & 70.46 & 51.28 & 79.51 & 52.95 & 86.37 & 72.94 & 11.2\\
         LXLv2 \cite{lxlv2}& C+R & 2025 & 47.81 & 49.30 & \underline{77.15} & 58.09 & - & - & - & - & 6.5\\
         Doracamom\cite{doracamom} & C+R & 2025 & 53.35 & 48.94 & 76.99 & 59.76 & - & - & - & - & 4.4\\
         HyDRa\cite{hydra} & C+R & 2025 & 52.83 & \underline{56.57} & 73.25 & 60.88 & 80.65 & 62.90 & 87.43 & 76.99 & \textit{N/A}\\
         MSSF\cite{mssf} & C+R & 2025 & 52.53 & 51.58 & 75.77 & 59.96 & 89.08 & \underline{66.78} & 88.10 & \underline{81.32} & 10.3 \\
    \hline
        SFGFusion\cite{sfgfusion} & C+R & 2025 & 48.30	& 43.60	& 75.54	& 55.76	& 79.10	& 53.63	& 88.60	& 73.77 & 6.4\\
        \rowcolor{lightblue} SFGFusion + M\textsuperscript{3} & C+R & 2025 & \underline{58.99} & 51.17 & \textbf{77.95} & \underline{62.70} & \underline{89.64} & 60.65 & \underline{88.91} & 79.73 & 5.2\\
         \textit{Improvement} & - & - & \textcolor{red}{\textit{+10.69}} & \textcolor{red}{\textit{+7.57}} & \textcolor{red}{\textit{+2.41}} & \textcolor{red}{\textit{+6.94}} & \textcolor{red}{\textit{+10.54}} & \textcolor{red}{\textit{+7.02}} & \textcolor{red}{\textit{+0.31}} & \textcolor{red}{\textit{+5.96}} & \textit{-1.2}\\
    \hline
        HGSFusion\cite{hgsfusion} & C+R & 2025 & 51.67 & 52.64 & 72.58 & 58.96 & 88.28 & 62.61 & 87.49 & 79.46 & 5.7 \\
        \rowcolor{lightblue} HGSFusion + M\textsuperscript{3} & C+R & 2025 & \textbf{61.52} & \textbf{62.87} & 76.04 & \textbf{66.81} &  \textbf{90.24} & \textbf{75.18} &  88.25 &  \textbf{84.56} & 4.4\\
         \textit{Improvement} & - & - & \textcolor{red}{\textit{+9.85}} & \textcolor{red}{\textit{+10.23}} & \textcolor{red}{\textit{+3.46}} & \textcolor{red}{\textit{+7.85}} & \textcolor{red}{\textit{+1.96}} & \textcolor{red}{\textit{+12.57}} & \textcolor{red}{\textit{+0.76}} & \textcolor{red}{\textit{+5.10}} & \textit{-1.3}\\
    
    \toprule[1.5pt]
    \end{tabular}
    \begin{tablenotes}    
    \footnotesize           
    \item[\textasteriskcentered]Region of Interest refers to the driving corridor near the ego-vehicle, defined in the camera coordinate as $D_{RoI} = \{ (x,y,z) \mid -4m < x < 4m, z < 25m \}$.
    \end{tablenotes}     
    \end{threeparttable}
\end{table*}

\subsection{Implementation Details}
\textbf{Hyper-parameter settings:}
The hyper-parameters used in our experiments follow the specifications provided by each dataset. For the VoD dataset, radar points are confined within a spatial range of 0m to 51.2m along the x-axis, -25.6m to 25.6m along the y-axis, and -3m to 2m along the z-axis. In the TJ4DRadSet dataset, the radar point cloud covers 0m to 69.12m in the x-axis, -39.68m to 39.68m in the y-axis, and -4m to 2m in the z-axis.

Moreover, predefined anchor boxes are applied in the detection head to facilitate 3D bounding box prediction. For the VoD dataset, the anchor sizes for cars, pedestrians, and cyclists are set to (3.9m, 1.6m, 1.56m), (0.8m, 0.6m, 1.73m), and (1.76m, 0.6m, 1.73m), respectively. For the TJ4DRadSet dataset, the anchor sizes for cars, pedestrians, cyclists, and trucks are set to (4.56m, 1.84m, 1.7m), (0.8m, 0.6m, 1.69m), (1.77m, 0.78m, 1.6m), and (10.76m, 2.66m, 3.47m), respectively.

\textbf{Training details:}
Our method is implemented on the OpenPCDet \cite{openpcdet2020} framework and trained on a single NVIDIA GeForce RTX 3090 GPU. M\textsuperscript{3}Detection leverages five frames of multi-modal data for multi-frame feature aggregation. During training, intermediate features generated from the baseline detector and corresponding tracking results are precomputed and incorporated into the training data to accelerate the model optimization. During inference, historical intermediate features are stored in the memory bank, while baseline feature extraction for the current frame and all modules of M\textsuperscript{3}Detection are executed online, ensuring that runtime performance evaluation reflects the actual system behavior. Model optimization is performed using the AdamW optimizer with an initial learning rate of 0.0002, and the network is trained for 18 epochs.

\subsection{Experiment Results}

\subsubsection{Results on VoD Dataset}
To evaluate the performance of our proposed model, we conducted experiments on the VoD dataset \cite{vod} based on two baseline detectors and compared the results with existing methods in TABLE \ref{Tab:VoD Result}. 

\textbf{Baseline detector enhancement:}
It can be observed that our method significantly improves baseline performance across all categories by integrating multi-frame information. Moreover, consistent gains obtained with two different baseline detectors further validate the general applicability of our approach. Baseline results show relatively low AP for cars and pedestrians, mainly because many stationary instances in the VoD dataset are hard for radar to distinguish from background noise. By integrating multi-frame information, our method produces denser and more reliable feature representations, leading to substantial accuracy improvements. Especially for pedestrians, enhanced modeling of local-level features in the multi-frame fusion strengthens the representation of small-scale objects and improves detection performance. However, the improvement for cyclists is less pronounced, mainly because cyclist samples are relatively scarce and most are in motion, where Doppler velocity cues from radar already provide strong motion information in single-frame detection, leaving limited room for further gains from multi-frame fusion.

\textbf{Comparison with the SOTA methods:}
Experimental results demonstrate that M\textsuperscript{3}Detection consistently outperforms existing methods in almost all metrics. Compared with the SOTA method MSSF, our method achieves significant gains in mAP, with improvements of 6.85\% and 3.24\% in the Entire Annotated Area (EAA) and Region of Interest (RoI), respectively. In particular, for the car and pedestrian categories, which constitute the majority of the dataset and pose the greatest challenge for single-frame detectors, M\textsuperscript{3}Detection achieves an EAA AP improvement of 8.17\% over the leading method Doracamom for cars and 6.30\% over the leading method HyDRA for pedestrians. These performance gains in the entire annotated area indicate that multi-frame feature aggregation effectively mitigates the impact of object motion state on detection performance and leverages the ranging capability of 4D imaging radar to enhance recognition, particularly for distant objects. In contrast, the improvement for cyclists is limited due to ambiguous and inconsistently annotated bicycle-like objects in the VoD dataset, such as parked bikes, racks, and scooters, which introduce label noise during training and weaken the effectiveness of multi-frame feature fusion.

\begin{table}[t]
    \centering
    \caption{Performance comparison of M\textsuperscript{3}Detection with LiDAR-based methods on the VoD dataset, where L denotes LiDAR, R denotes 4D imaging radar, and C represents camera}
    \label{Tab: VoD  Compare With Lidar}
    \fontsize{8pt}{13pt}\selectfont
    \setlength{\tabcolsep}{0.92pt}
    \begin{tabular}{c|c|ccc|c|ccc|c}
    \bottomrule[1.3pt]
    \multirow{2}{*}{\centering Method} & \multirow{2}{*}{\centering Modality} & \multicolumn{4}{c|}{AP in the EAA (\%)} & \multicolumn{4}{c}{AP in the RoI (\%)} \\
    \cline{3-10}
         & & Car & Ped & Cyc & mAP & Car & Ped & Cyc & mAP\\ 
    \hline
        PointPillars\cite{pointpillars} & L & 59.11 & 37.71 & 64.49 & 53.77 & \textbf{92.35} & 48.02 & \textbf{89.08} & 76.48 \\
        Voxel Mamba\cite{voxel_mamba} & L & 67.28 & \textbf{65.34} & 74.38 & \underline{69.00} & \underline{90.67} & \textbf{78.14} & 86.40 & \textbf{85.07} \\
        MVX-Net\cite{mvx-net} & C+L & \textbf{78.25} & 51.76 & 52.26 & 60.76 & 90.62 & 68.16 & 72.03 & 76.94 \\
        InterFusion\cite{interfusion} & R+L & \underline{67.50} & \underline{63.21} & \textbf{78.79} & \textbf{69.83} & 88.11 & 74.80 & 87.50 & 83.47 \\
         \rowcolor{lightblue} SFGFusion + M\textsuperscript{3} & C+R & 58.99& 51.17 & \underline{77.95} & 62.70 & 89.64 & 60.65 & \underline{88.91} & 79.73 \\
         \rowcolor{lightblue} HGSFusion + M\textsuperscript{3} & C+R & 61.52 & 62.87 & 76.04 & 66.81 & 90.24 & \underline{75.18} &  88.25 &  \underline{84.56}\\
    \toprule[1.5pt]
    \end{tabular}
\end{table}

\textbf{Comparison with the LiDAR methods:}
As shown in TABLE \ref{Tab: VoD  Compare With Lidar}, our approach significantly narrows the performance gap with LiDAR-involved methods through multi-frame feature aggregation. In the RoI, HGSFusion + M\textsuperscript{3}Detection achieves an mAP of 84.56\%, surpassing the classical LiDAR-only method PointPillars, camera-LiDAR fusion method MVX-Net and the radar-LiDAR fusion method InterFusion. In the EAA, although our method is constrained by the inherent resolution and noise limitations of 4D imaging radar, it benefits from multi-frame multi-level feature fusion, achieving comparable performance to LiDAR-involved methods and exceeding several of them in particular categories. Overall, these findings demonstrate that multi-frame integration effectively mitigates the shortcomings of low-cost sensors, enabling camera-4D radar fusion to rival LiDAR in specific categories and scenarios.

\begin{figure*}[t]
\centering{\includegraphics[width= \textwidth , height=\textheight, keepaspectratio]{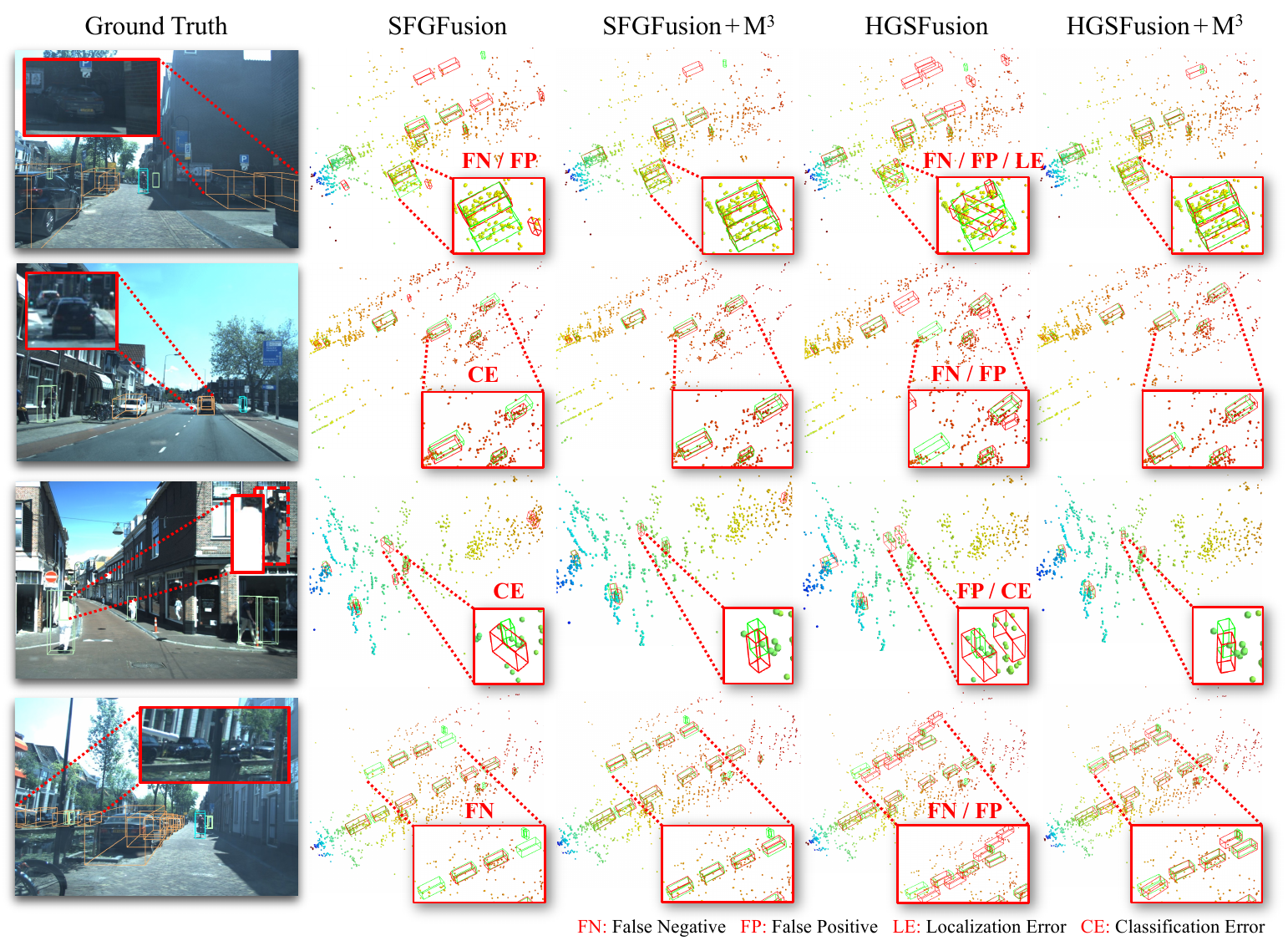}}
\caption{\textbf{Visualization of M\textsuperscript{3}Detection results on the VoD dataset\cite{vod}.} Each row illustrates a representative scene with both image and 3D point cloud views. In the image view, ground-truth objects of different categories are distinguished using different colors, while in the point cloud view, green and red bounding boxes denote ground truth and predictions, respectively. From left to right, each column corresponds to Ground Truth, SFGFusion, SFGFusion+M\textsuperscript{3}, HGSFusion, and HGSFusion+M\textsuperscript{3}. FN denotes false negative, FP denotes false positive, LE denotes localization error, and CE denotes classification error. By leveraging multi-frame temporal information, M\textsuperscript{3}Detection achieves superior detection under challenging conditions, including truncation, occlusion, complex backgrounds, and distant objects, consistently improving baseline detectors.}
\label{Fig: VoD Result}
\end{figure*}

\begin{table*}[h]
    \centering
    \caption{Comparison of detectors' results on the TJ4DRadSet dataset \cite{tj4dradset}, where R denotes 4D imaging radar and C represents camera. Bold and underlined values denote the best and second-best performance, respectively}
    \fontsize{10pt}{13pt}\selectfont
    \setlength{\tabcolsep}{4.3pt}
    \label{Tab: TJ4DRadset Result}
    \begin{tabular}{c|c|c|cccc|c|cccc|c}
    \bottomrule[1.5pt]
    \multirow{2}{*}{\centering Method} & \multirow{2}{*}{\centering Modality} & \multirow{2}{*}{\centering Year}& \multicolumn{5}{c|}{3D (\%)} & \multicolumn{5}{c}{BEV (\%)} \\
    \cline{4-13}
         & & & Car & Ped & Cyc & Tru & mAP & Car & Ped & Cyc & Tru & mAP \\
    \hline
        ImVoxelNet\cite{imvoxelnet} & C & 2022 & 22.55 & 13.73 & 9.67 & 13.87 & 14.96 & 26.10 & 14.21 & 10.99 & 17.18 & 17.12 \\
    \hline
        SECOND\cite{second} & R & 2018 & 18.18 & 24.43 & 32.36 & 14.76 & 22.43 & 36.02 & 28.58 & 39.75 & 19.35 & 30.93 \\
        PointPillars\cite{pointpillars} & R & 2019 & 21.26 & 28.33 & 52.47 & 11.18 & 28.31 & 38.34 & 32.26 & 56.11 & 18.19 & 36.23 \\
        Part-A$^2$\cite{part-A} & R & 2020 & 18.65 & 23.28 & 44.14 & 9.63 & 23.92 & 29.95 & 24.31 & 49.08 & 15.05 & 29.60 \\
        CenterPoint\cite{centerpoint} & R & 2021 & 22.03 & 25.02 & 53.32 & 15.92 & 29.07 & 33.03 & 27.87 & 58.74 & 25.09 & 36.18 \\
        RPFA-Net\cite{RPFA-Net} & R & 2021 & 26.89 & 27.36 & 50.95 & 14.46 & 29.91 & 42.89 & 29.81 & 57.09 & 25.98 & 38.94 \\
        PillarNeXt\cite{pillarnext} & R & 2023 & 22.33 & 23.48 & 53.01 & 17.99 & 29.20 & 36.84 & 25.17 & 57.07 & 23.76 & 35.71 \\
        RadarPillarNet\cite{RCFusion} & R & 2023 & 28.45 & 26.24 & 51.57 & 15.20 & 30.37 & 45.72 & 29.19 & 56.89 & 25.17 & 39.24 \\
        SMURF\cite{smurf} & R & 2024 & 28.47 & 26.22 & 54.61 & 22.64 & 32.99 & 43.13  & 29.19 & 58.81 & 32.80 & 40.98 \\
    \hline
        MVX-Net\cite{mvx-net} & C+R & 2019 & 22.28 & 19.57 & 50.70 & 11.21 & 25.94 & 37.46 & 22.70 & 54.69 & 18.07 & 33.23 \\
        PointAugmenting\cite{pointaugmenting} & C+R & 2021 & 22.63 & 26.23 & 53.52 & 13.37 & 28.94 & 43.42 & 29.65 & 59.21 & 23.88 & 39.04 \\
        Focals Conv\cite{focal} & C+R & 2022 & 12.24 & \underline{31.80} & 54.01 & 6.66 & 26.18 & 22.52 & \underline{35.33} & 59.37 & 10.30 & 31.88 \\
        FUTR3D\cite{futr3d} & C+R & 2023 & - & - & - & - & 32.42 & - & - & - & - & 37.51 \\
        BEVFusion\cite{bevfusion} & C+R & 2023 & - & - & - & - & 32.71 & - & - & - & - & 41.12 \\
        RCFusion\cite{RCFusion} & C+R & 2023 & 29.72 & 27.17 & 54.93 & 23.56 & 33.85 & 40.89 & 30.95 & 58.30 & 28.92 & 39.76 \\
        LXL\cite{LXL} & C+R & 2024 & - & - & - & - & 36.32 & - & - & - & - & 41.20 \\
        SGDet3D\cite{sgdet3d} & C+R & 2024 & \textbf{59.43} & 26.57& 51.30 & 30.00 & 41.82 & \textbf{66.38} & 29.18 & 53.72 & 39.36 & 47.16 \\
        LXLv2\cite{lxlv2} & C+R & 2025 & - & - & - & - & 37.32 & - & - & - & - & 42.35 \\
        MSSF\cite{mssf} & C+R & 2025 & 45.18 & \textbf{33.61} & 55.88 & 17.20 & 37.97 & 56.25 & \textbf{36.53} & 58.70 & 20.97 & 43.11\\
    \hline 
        SFGFusion\cite{sfgfusion} & C+R & 2025 & 33.05 & 27.01 & 55.12 & 27.10 & 35.57 & 46.63 & 30.04 & 59.55 & 42.41 & 44.66 \\
        \rowcolor{lightblue} SFGFusion + M\textsuperscript{3} & C+R & 2025 & 45.37 & 28.20 & \underline{64.00} & \textbf{35.45} & \underline{43.25} & 58.17 & 31.65 & \underline{66.82} & \textbf{47.43} & \underline{51.02} \\
         \textit{Improvement} & - & - & \textcolor{red}{\textit{+}12.32} & \textcolor{red}{\textit{+1.19}} & \textcolor{red}{\textit{+8.88}} & \textcolor{red}{\textit{+8.35}} & \textcolor{red}{\textit{+7.68}} & \textcolor{red}{\textit{+11.54}} & \textcolor{red}{\textit{+1.61}} & \textcolor{red}{\textit{+7.27}} & \textcolor{red}{\textit{+5.02}} & \textcolor{red}{\textit{+6.36}}\\
    \hline
        HGSFusion\cite{hgsfusion} & C+R & 2025 & - & - & - & - & 37.21 & & - & - & - & 43.23  \\
        \rowcolor{lightblue} HGSFusion + M\textsuperscript{3} & C+R & 2025 & \underline{51.94} & 25.70 & \textbf{69.25} & \underline{32.01} &  \textbf{44.73} & \underline{59.13} & 29.58 & \textbf{71.46} & \underline{44.72} & \textbf{51.22}\\
         \textit{Improvement} & - & - & - & - & - & - & \textcolor{red}{\textit{+7.52}} & - & - & - & - & \textcolor{red}{\textit{+7.99}}\\
    
    \toprule[1.5pt]
    \end{tabular}    
\end{table*}

\textbf{Model size and inference speed:}
TABLE \ref{Tab:VoD Result} reports the computational efficiency of different models. Compared with the baseline detector, M\textsuperscript{3}Detection achieves significant performance improvements while maintaining reasonable runtime efficiency. By aggregating historical intermediate features from the baseline detector, our method avoids redundant re-extraction of trajectory point clouds, thereby controlling computational cost effectively. In terms of model size, M\textsuperscript{3}Detection contains only 31.3M parameters, significantly fewer than the single-frame baselines SFGFusion (191.9M) and HGSFusion (246.5M). Although multi-frame aggregation inevitably introduces additional inference overhead, the incurred slowdown remains within acceptable bounds and is well justified by the performance improvement achieved. Overall, the proposed method achieves notable accuracy improvements while preserving a lightweight design, and it can be flexibly integrated as a general component into a wide range of single-frame detectors.

\textbf{Visualization:} 
\cref{Fig: VoD Result} presents the qualitative results of M\textsuperscript{3}Detection on representative scenes from the VoD dataset. Leveraging multi-modal sequential inputs, M\textsuperscript{3}Detection achieves stable and accurate performance under challenging conditions such as truncation, occlusion, complex backgrounds, and distant objects, significantly enhancing the recognition and localization capabilities of baseline detectors. In the first-row scene, the vehicle on the right side of the image is truncated and nearly invisible. Although radar reflections provide partial point cloud information, single-frame baseline detectors struggle to correctly identify and localize the object due to radar point sparsity and the absence of image guidance. In contrast, M\textsuperscript{3}Detection aggregates temporal radar features across frames to recover the object and achieve reliable detection. In the third-row scene, a pedestrian is completely occluded by another, making single-frame detection infeasible. By leveraging temporal cues from historical features, M\textsuperscript{3}Detection accurately localizes the occluded pedestrian in the current frame, demonstrating superior spatiotemporal reasoning and robustness. Overall, M\textsuperscript{3}Detection effectively integrates multi-modal and temporal information, yielding more consistent and reliable 3D object detection results.

\subsubsection{Results on TJ4DRadSet Dataset}
To further evaluate the performance of our proposed model, we conducted additional tests on the TJ4DRadset dataset \cite{tj4dradset} based on two baseline detectors as shown in TABLE \ref{Tab: TJ4DRadset Result}.

\textbf{Baseline detector enhancement:}
Compared with the VoD dataset, the TJ4DRadSet presents more complex scenarios, such as nighttime, glare, and camera defocus, which pose greater challenges for detection and thus provide a more rigorous test of multi-frame feature modeling. As HGSFusion does not report category-wise AP values in TJ4DRadSet, we compare only the overall mAP improvements for this baseline. As shown in TABLE \ref{Tab: TJ4DRadset Result}, M\textsuperscript{3}Detection leverages temporal aggregation to overcome the incomplete representations of single-frame detection, leading to notable performance gains in both 3D and BEV metrics for the two baselines. At the category level, substantial improvements are observed for cars and cyclists, which dominate the dataset in terms of frequency. In addition, TJ4DRadSet introduces a truck class characterized by large intra-class scale variation, further increasing detection difficulty. Benefiting from multi-level feature aggregation, M\textsuperscript{3}Detection captures both global and local cues, thus adapting well to objects of diverse scales. For pedestrians, however, the scarcity of training samples and weak radar reflections constrain the performance gains from multi-frame feature fusion.

\begin{figure*}[t]
\centering{\includegraphics[width=\textwidth , height=\textheight, keepaspectratio]{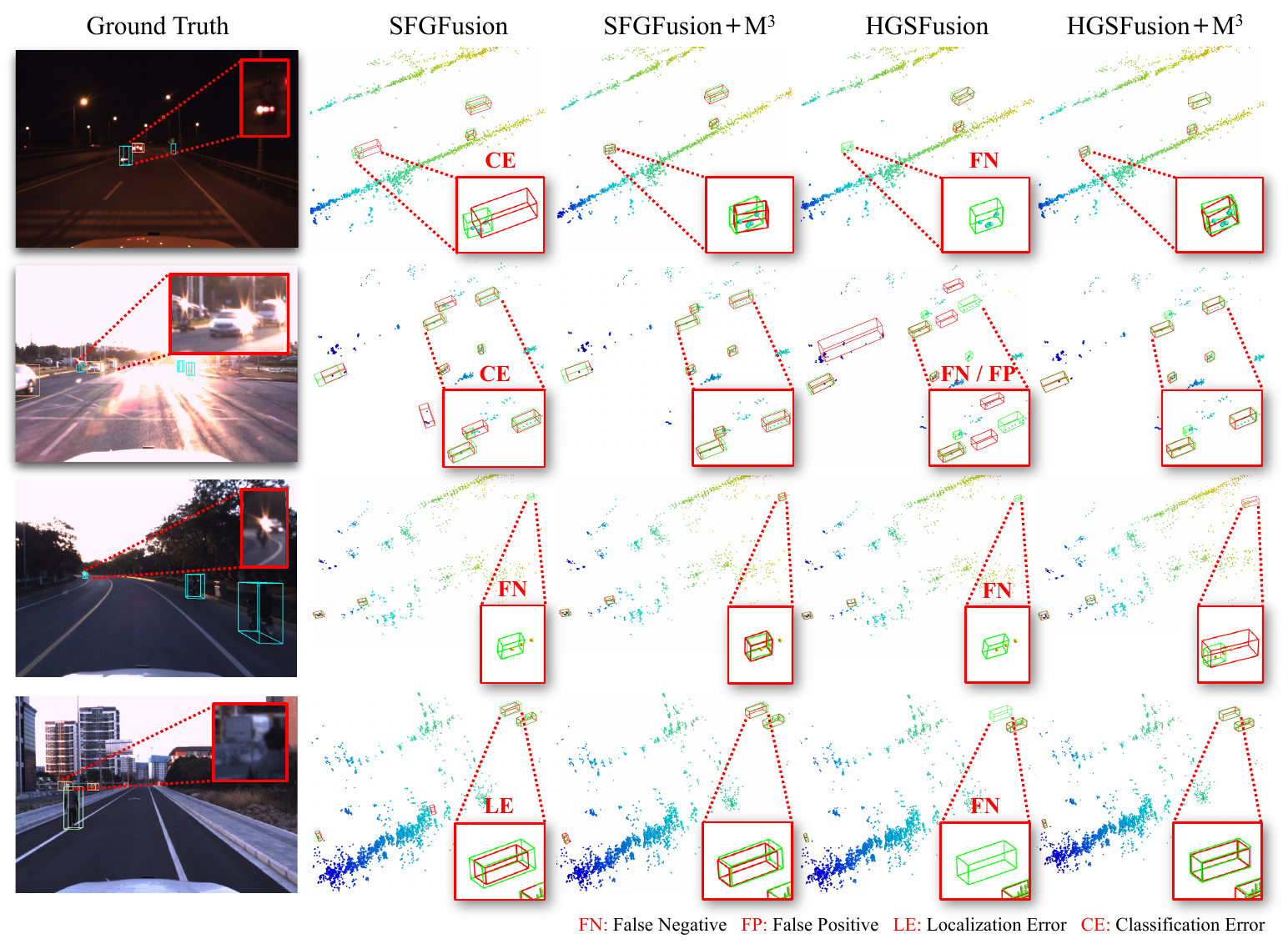}}
\caption{\textbf{Visualization of M\textsuperscript{3}Detection results on the TJ4DRadSet dataset\cite{tj4dradset}.} Each row illustrates a representative scene with both image and 3D point cloud views. In the image view, ground-truth objects of different categories are distinguished using different colors, while in the point cloud view, green and red bounding boxes denote ground truth and predictions, respectively. From left to right, each column corresponds to Ground Truth, SFGFusion, SFGFusion+M\textsuperscript{3}, HGSFusion, and HGSFusion+M\textsuperscript{3}. FN denotes false negative, FP denotes false positive, LE denotes localization error, and CE denotes classification error. By leveraging multi-frame temporal information, M\textsuperscript{3}Detection achieves superior detection under challenging conditions, including extreme illumination, truncation, occlusion, complex backgrounds, and distant objects, consistently improving baseline detectors.}
\label{Fig: TJ4D Result}
\end{figure*}

\textbf{Comparison with the SOTA methods:}
The results on TJ4DRadSet show that M\textsuperscript{3}Detection consistently outperforms existing methods across most categories. Specifically, compared with the SOTA method SGDet3D, M\textsuperscript{3}Detection achieves notable improvements of 2.91\% and 4.06\% in 3D mAP and BEV mAP, respectively, confirming the effectiveness of the proposed multi-frame detection framework. Furthermore, the most striking gains are observed in the cyclist and truck categories, where M\textsuperscript{3}Detection achieves a 3D AP improvement of 13.37\% over the leading method MSSF for cyclists and 15.45\% over the leading method SGDet3D for trucks. These findings indicate that our method can effectively adapt to objects of different scales: the local-level feature aggregation module captures fine-grained geometric details that mitigate sparsity and incomplete information of single-frame features for small objects, while the global-level feature aggregation module provides semantic representations that enhance robustness and stability when handling large objects. For the car and pedestrian categories, M\textsuperscript{3}Detection delivers significant improvements over the baseline detectors; however, due to the inherently weaker baseline performance, its advantage over competing methods is less pronounced.

\begin{table}[t]
    \centering
    \caption{Evaluation of M\textsuperscript{3}Detection across lighting conditions}
    \label{Tab: Performance across lighting conditions}
    \fontsize{8pt}{13pt}\selectfont
    \setlength{\tabcolsep}{3.5pt}
    \begin{tabular}{c|ccc|ccc}
    \bottomrule[1.5pt]
    \multirow{2}{*}{\centering Method}  & \multicolumn{3}{c|}{3D mAP (\%)} & \multicolumn{3}{c}{BEV mAP (\%)} \\
    \cline{2-7}
         & Dark & Normal & Shiny & Dark & Normal & Shiny \\ 
    \hline
        SFGFusion & 19.95 & 26.96 & 28.46 & 24.29 & 31.22 & 42.88 \\
        \rowcolor{lightblue} SFGFusion + M\textsuperscript{3} & 33.50 & 30.76 & 36.44 & 38.66 & 35.83 & 48.51 \\
        \textit{Improvement} & \textcolor{red}{\textit{+13.55}} & \textcolor{red}{\textit{+3.80}} & \textcolor{red}{\textit{+7.98}} & \textcolor{red}{\textit{+14.37}} & \textcolor{red}{\textit{+4.61}} & \textcolor{red}{\textit{+5.63}} \\
        HGSFusion & 15.68 & 35.82 & 25.28 & 19.73 & 42.05 & 31.83 \\
        \rowcolor{lightblue} HGSFusion + M\textsuperscript{3} & 22.86 & 36.62 & 44.28 & 26.95 & 42.21 & 50.61\\
        \textit{Improvement} & \textcolor{red}{\textit{+7.18}} & \textcolor{red}{\textit{+0.80}} & \textcolor{red}{\textit{+19.00}} & \textcolor{red}{\textit{+7.22}} & \textcolor{red}{\textit{+0.16}} & \textcolor{red}{\textit{+18.78}} \\   
    \toprule[1.5pt]
    \end{tabular}
\end{table}

\begin{table}[t]
    \centering
    \caption{Evaluation of M\textsuperscript{3}Detection across object distances}
    \label{Tab: Performance across object distances}
    \fontsize{8pt}{13pt}\selectfont
    \setlength{\tabcolsep}{3.5pt}
    \begin{tabular}{c|ccc|ccc}
    \bottomrule[1.5pt]
    \multirow{2}{*}{\centering Method}  & \multicolumn{3}{c|}{3D mAP (\%)} & \multicolumn{3}{c}{BEV mAP (\%)} \\
    \cline{2-7}
         & 0-25m & 25-50m & 50-70m & 0-25m & 25-50m & 50-70m \\ 
    \hline
        SFGFusion & 47.31 & 31.34 & 13.41 & 54.36 & 40.54 & 21.22 \\
        \rowcolor{lightblue} SFGFusion + M\textsuperscript{3} & 53.78 & 37.10 & 23.09 & 59.87 & 44.46 & 31.89 \\
        \textit{Improvement} & \textcolor{red}{\textit{+6.47}} & \textcolor{red}{\textit{+5.76}} & \textcolor{red}{\textit{+9.68}} & \textcolor{red}{\textit{+5.51}} & \textcolor{red}{\textit{+3.92}} & \textcolor{red}{\textit{+10.67}} \\
        HGSFusion & 54.54 & 30.16 & 18.23 & 60.83 & 36.74 & 22.11 \\
        \rowcolor{lightblue} HGSFusion + M\textsuperscript{3} & 54.98 & 41.95 & 22.82 & 60.53 & 47.33 & 28.34 \\
        \textit{Improvement} & \textcolor{red}{\textit{+0.44}} & \textcolor{red}{\textit{+11.79}} & \textcolor{red}{\textit{+4.59}} & \textcolor{red}{\textit{-0.30}} & \textcolor{red}{\textit{+10.59}} & \textcolor{red}{\textit{+6.23}} \\
    \toprule[1.5pt]
    \end{tabular}
\end{table}

\textbf{Performance across lighting conditions:}
To evaluate the robustness of our method under varying lighting conditions, we follow the data partitioning protocol of HGSFusion \cite{hgsfusion} and divide the TJ4DRadSet dataset into three subsets: Dark, Normal, and Shiny. As shown in TABLE \ref{Tab: Performance across lighting conditions}, M\textsuperscript{3}Detection consistently improves detection performance across all lighting conditions, demonstrating its effectiveness in diverse environments. In particular, performance improvement is most notable under dark and shiny conditions, where degraded image quality leads to noisy or missing texture features in single-frame detectors. By aggregating multi-frame features and modeling temporal context, M\textsuperscript{3}Detection effectively stabilizes feature representation, leading to higher accuracy and robustness. Specifically, under dark and shiny conditions, the 3D mAP of SFGFusion increases by 13.55\% and 7.98\%, whereas HGSFusion achieves gains of 7.18\% and 19.00\%, respectively. Conversely, under normal lighting, where visual quality is stable and image cues are sufficiently captured by the single-frame baselines, the multi-frame enhancement becomes less pronounced yet still consistent. These findings indicate that M\textsuperscript{3}Detection provides substantial robustness under extreme illumination while maintaining steady improvements under standard conditions.

\textbf{Performance across object distances:} 
We evaluated the detection performance across different object distance ranges, as shown in TABLE \ref{Tab: Performance across object distances}. M\textsuperscript{3}Detection consistently enhances the accuracy of baseline detectors across almost all distance ranges, benefiting from spatial feature aggregation and temporal cross-frame reasoning. With increasing distance, single-frame detectors suffer from diminished visual detail and sparser radar points, resulting in significant performance decline. In contrast, M\textsuperscript{3}Detection leverages multi-frame information aggregation to compensate for missing observations and extract more stable object representations, effectively mitigating the performance degradation. Specifically, for the most challenging 50-70m distance range, M\textsuperscript{3}Detection improves the 3D mAP by 9.68\% and 4.59\% over the two baseline detectors. Conversely, the BEV mAP improvement in the 0-25m near-field range is relatively limited, primarily because the feature refinement during multi-frame aggregation may introduce minor spatial shifts in high-density scenarios, slightly affecting BEV localization. Overall, M\textsuperscript{3}Detection maintains stable gains across distances, with particularly notable improvements for mid- to long-range objects, demonstrating strong robustness and generalization.

\begin{table*}[t]
    \centering
    \caption{Ablation studies on key components. Experiments are conducted on the VoD dataset\cite{vod}. Bold and underlined values denote the best and second-best performance, respectively}
    \label{Tab: Ablation on key components}
    \begin{threeparttable}
    \fontsize{10pt}{13pt}\selectfont
    \setlength{\tabcolsep}{8pt}
    \begin{tabular}{c|c|c|c|ccc|c|ccc|c}
    \bottomrule[1.5pt]    
    \multirow{2}{*}{\centering Baseline} & \multicolumn{3}{c|}{Component} &  \multicolumn{4}{c|}{AP in the Entire Annotated Area (\%)} & \multicolumn{4}{c}{AP in the Region of Interest (\%)} \\
    \cline{2-12}
         & MSTR & GOA & LGA & Car & Ped & Cyc & mAP & Car & Ped & Cyc & mAP \\
    \hline
        \checkmark &  &  &  & 51.67 & 52.64 & 72.58 & 58.96 & 88.28 & 62.61 & 87.49 & 79.46 \\
        
        \checkmark & \checkmark & \checkmark &  & 55.72	& 56.55	& 74.83	& 62.37	& \underline{88.58}	& 72.79	& \underline{88.07}	& 83.15 \\
        \checkmark & \checkmark &  & \checkmark & \underline{56.00} & \underline{57.90} & \underline{75.28} & \underline{63.06} & 88.55 & \underline{73.33} & 88.06 & \underline{83.32} \\
        \rowcolor{lightblue} \checkmark & \checkmark & \checkmark & \checkmark & \textbf{61.52} & \textbf{62.87} & \textbf{76.04} & \textbf{66.81} & \textbf{90.24} & \textbf{75.18} &  \textbf{88.25} &  \textbf{84.56}\\
    \toprule[1.5pt]    
    \end{tabular}       
    \end{threeparttable}
\end{table*}

\begin{table}[t]
    \centering
    \caption{Ablation studies on global-level feature aggregation. Experiments are conducted on the VoD dataset\cite{vod}}
    \label{Tab: Ablation on global-level GOA}
    \begin{threeparttable}
    \fontsize{10pt}{13pt}\selectfont
    \setlength{\tabcolsep}{4pt}
    \begin{tabular}{c|>{\centering\arraybackslash}p{1.2cm}|>{\centering\arraybackslash}p{1.2cm}|c|c}
    \bottomrule[1.5pt]    
    \multirow{2}{*}{\centering Hypothesis} & \multicolumn{2}{c|}{Component} & \multirow{2}{*}{\centering mAP$_{\text{EAA}}$(\%)} & \multirow{2}{*}{mAP$_{\text{RoI}}$(\%)}\\
    \cline{2-3}
         & PE & RG &  & \\
    \hline
        Single &  &  & 59.04 & 81.06 \\
        Multi &  &  & 60.81 & 80.58 \\
        Multi & \checkmark &  & 61.29 & 81.39 \\
        Multi &  & \checkmark & 61.52 & 82.45 \\
        \rowcolor{lightblue} Multi & \checkmark & \checkmark & \textbf{62.37} & \textbf{83.15}\\
    \toprule[1.5pt]    
    \end{tabular}
    \begin{tablenotes}    
    \footnotesize              
    \item[*]RG: radar information guidance, PE: position embedding.
    \end{tablenotes}            
    \end{threeparttable}
\end{table}

\textbf{Visualization:}
\cref{Fig: TJ4D Result} showcases the visualization results of M\textsuperscript{3}Detection on representative scenes from the TJ4DRadSet dataset. By aggregating multi-modal temporal information, our method effectively mitigates challenges such as extreme illumination, truncation, occlusion, complex backgrounds, and distant objects, maintaining precise and stable detection under sensor degradation. In the first three rows, low-light or strong glare conditions severely impair or disable the camera in the current frame, making single-frame detection insufficient. M\textsuperscript{3}Detection exploits historical image and radar features to aggregate cross-frame information and perform spatiotemporal reasoning, enabling accurate capture and localization of objects. In the fourth-row scene, sparse or missing radar returns for distant objects prevent single-frame detectors from reliably estimating 3D position and size. By integrating multi-frame historical features and temporal context, M\textsuperscript{3}Detection effectively compensates for missing data and achieves robust detection. Overall, the framework enables accurate object detection in complex scenarios despite degraded or missing sensor inputs, effectively enhancing the robustness of baseline detectors.

\subsection{Ablation Experiments}
To evaluate the impact of individual modules on the overall detection framework, we conduct ablation experiments on the VoD dataset, with HGSFusion serving as the single-frame baseline detector.

\textbf{Ablation on key components:}
This section evaluates the contributions of different feature aggregation components through ablation studies, where aggregated single-frame features are further refined in the trajectory-level multi-frame spatiotemporal reasoning module (MSTR), boosting overall detection performance. As shown in TABLE \ref{Tab: Ablation on key components}, both the global-level inter-object feature aggregation module (GOA) and the local-level inter-grid feature aggregation (LGA) provide significant performance gains. Specifically, GOA performs inter-object aggregation on global intermediate features containing high-level semantic information, effectively capturing potential objects while mitigating feature loss, thereby improving overall model performance. LGA focuses on aggregating local intermediate features that encode fine-grained object information, preserving and reinforcing structural details, particularly enhancing the detection of small objects such as pedestrians and cyclists. By integrating GOA and LGA for global and local feature aggregation with multi-frame spatiotemporal reasoning via the MSTR module, the overall architecture enables cross-frame information fusion, leading to substantial improvements in detection performance.

\textbf{Ablation on global-level inter-object feature aggregation:}
To verify the effectiveness of the GOA, we performed ablation studies on its key components using only global-level features without local-level representations. Specifically, we compared the effects of direct feature aggregation under reference trajectories (single-hypothesis) and candidate proposals (multi-hypothesis) tracking results and analyzed the influence of the radar information guidance (RG) and position embedding (PE) modules on feature aggregation. As shown in TABLE \ref{Tab: Ablation on global-level GOA}, each component contributes to performance improvement, while their combination achieves the best results. In particular, the RG module introduces spatial cues from radar to suppress noise from invalid candidate proposals, and the PE module encodes positional information to enhance object representation and feature alignment. Under the same multi-hypothesis setting, incorporating RG and PE leads to 1.56\% and 2.57\% 3D mAP improvements in the EAA and RoI regions, respectively. Notably, without the PE and RG module, multi-hypothesis aggregation performs slightly worse than single-hypothesis aggregation in dense or near-range scenes within the RoI, as redundant candidate proposals may lead to feature confusion and incorrect aggregation. With the introduction of PE and RG, the model effectively filters valid object features while retaining diverse candidates, thereby improving the discriminability and stability of the aggregated features and mitigating the adverse effects of track loss and mistracking.

\textbf{Ablation on local-level inter-grid feature aggregation:}
We conducted ablation studies to evaluate the effectiveness of individual components within the LGA without global-level feature aggregation. Specifically, the experiments analyzed the contributions of crop region expansion (CRE) and cross-level deformable attention (CDA), highlighting their roles in enhancing local-level feature aggregation. As shown in TABLE \ref{Tab: Ablation on local-level LGA}, integrating feature aggregation based on reference trajectories with CRE and CDA notably improves 3D detection accuracy. Specifically, the CRE module expands the local feature extraction region to capture more complete object appearances under object motion or tracking errors, improving feature alignment and region coverage. On this basis, the CDA module leverages global-level query features to guide selective aggregation of key local features through cross-level deformable attention, enhancing feature representation while enabling global-local interaction. The combined incorporation of CRE and CDA results in 3D mAP gains of 3.11\% and 2.58\% in the EAA and RoI regions, respectively. These results indicate that the two components complement each other within the LGA, jointly enhancing local feature aggregation and improving 3D detection performance.

\begin{table}[t]
    \centering
    \caption{Ablation studies on local-level feature aggregation. Experiments are conducted on the VoD dataset\cite{vod}}
    \label{Tab: Ablation on local-level LGA}
    \begin{threeparttable}
    \fontsize{10pt}{13pt}\selectfont
    \setlength{\tabcolsep}{8pt}
    \begin{tabular}{>{\centering\arraybackslash}p{1.3cm}|>{\centering\arraybackslash}p{1.3cm}|c|c}
    \bottomrule[1.5pt]    
    \multicolumn{2}{c|}{Component} & \multirow{2}{*}{\centering mAP$_{\text{EAA}}$(\%)} & \multirow{2}{*}{mAP$_{\text{RoI}}$(\%)}\\
    \cline{1-2}
         CRE & CDA &  & \\
    \hline
         &  & 59.95 & 80.74\\
         \checkmark &  & 61.58 & 81.11 \\
        \rowcolor{lightblue} \checkmark & \checkmark & \textbf{63.06} & \textbf{83.32}\\
    \toprule[1.5pt]    
    \end{tabular}
    \begin{tablenotes}    
    \footnotesize              
    \item[*]CRE: crop region expansion, CDA: cross-level deformable attention.
    \end{tablenotes}            
    \end{threeparttable}
\end{table}

\begin{table}[t]
    \centering
    \caption{Ablation studies on input sequence length. Experiments are conducted on the VoD dataset\cite{vod}}
    \label{Tab: Ablation on input sequence length}
    \fontsize{9pt}{13pt}\selectfont
    \setlength{\tabcolsep}{3.8pt}
    \begin{tabular}{c|c|c|c|c}
    \bottomrule[1.5pt]
    Length & mAP$_{\text{EAA}}$(\%) & \textit{Improvement} & mAP$_{\text{RoI}}$(\%) & \textit{Improvement}\\
    \hline
        1 frame  & 58.96 & - & 79.46 & - \\
        2 frames & 61.48 & \textcolor{red}{\textit{+2.52}} & 81.58 & \textcolor{red}{\textit{+2.12}} \\
        3 frames & 62.57 & \textcolor{red}{\textit{+3.61}} & 83.77 & \textcolor{red}{\textit{+4.31}} \\
        4 frames & 63.89 & \textcolor{red}{\textit{+4.93}} & 84.10 & \textcolor{red}{\textit{+4.64}} \\
        \rowcolor{lightblue} 5 frames & \textbf{66.81} & \textcolor{red}{\textit{+7.85}} & \textbf{84.56} & \textcolor{red}{\textit{+5.10}} \\
    \toprule[1.5pt]
    \end{tabular}
\end{table}

\textbf{Ablation on input sequence length:}
In this section, we evaluate the impact of varying input sequence length on the performance of M\textsuperscript{3}Detection. While single-frame inputs suffer from occlusion, truncation, and modality limitations, M\textsuperscript{3}Detection takes advantage of multi-modal sequential data, leading to consistent accuracy improvements as the number of frames increases. The results in TABLE \ref{Tab: Ablation on input sequence length} demonstrate that longer sequences allow more comprehensive cross-frame feature aggregation and reinforce temporal consistency in object representation. Furthermore, the integration of multi-frame spatiotemporal reasoning at the trajectory-level enables the model to continuously improve detection accuracy as the input sequence length increases, without suffering from performance degradation caused by cross-object interference. While sequences longer than five frames may offer additional performance gains, we adopt five frames as the default input length for M\textsuperscript{3}Detection, achieving a balanced trade-off between performance and efficiency.

\section{Conclusion}\label{Conclusion}
In this paper, we present M\textsuperscript{3}Detection, a unified multi-frame 3D object detection framework that performs multi-level feature fusion on multi-modal sequential data from camera and 4D imaging radar. M\textsuperscript{3}Detection effectively addresses the challenges of redundant feature extraction and incomplete single-frame perception by leveraging historical intermediate features and integrating temporal reasoning across multiple frames. The framework utilizes a multi-modal baseline detector to extract intermediate features and produce initial detection results, while the specially designed tracker generates reference trajectories and candidate proposals. Building upon this, the global-level inter-object feature aggregation (GOA) module leverages radar cues to align and integrate global features from multiple candidate proposal positions, while the local-level inter-grid feature aggregation (LGA) module utilizes crop region expansion and cross-level deformable attention to aggregate fine-grained local features based on reference trajectories. The aggregated features are further refined through the trajectory-level multi-frame spatiotemporal reasoning (MSTR) module, enabling consistent and robust 3D perception across time.

Extensive experiments on the VoD and TJ4DRadSet datasets demonstrate that M\textsuperscript{3}Detection achieves state-of-the-art performance and consistently improves upon different baseline detectors. These results underscore the generality and adaptability of our framework, highlighting its potential to serve as a plug-and-play enhancement module for accurate and robust 3D object detection with camera-4D imaging radar fusion. In future work, we plan to extend M\textsuperscript{3}Detection toward dynamic scene understanding tasks such as 3D occupancy prediction and motion forecasting, further enhancing temporal modeling and environmental perception.


\bibliography{IEEEabrv}

\begin{thebibliography}{10}
\providecommand{\url}[1]{#1}
\csname url@samestyle\endcsname
\providecommand{\newblock}{\relax}
\providecommand{\bibinfo}[2]{#2}
\providecommand{\BIBentrySTDinterwordspacing}{\spaceskip=0pt\relax}
\providecommand{\BIBentryALTinterwordstretchfactor}{4}
\providecommand{\BIBentryALTinterwordspacing}{\spaceskip=\fontdimen2\font plus
\BIBentryALTinterwordstretchfactor\fontdimen3\font minus \fontdimen4\font\relax}
\providecommand{\BIBforeignlanguage}[2]{{%
\expandafter\ifx\csname l@#1\endcsname\relax
\typeout{** WARNING: IEEEtran.bst: No hyphenation pattern has been}%
\typeout{** loaded for the language `#1'. Using the pattern for}%
\typeout{** the default language instead.}%
\else
\language=\csname l@#1\endcsname
\fi
#2}}
\providecommand{\BIBdecl}{\relax}
\BIBdecl

\bibitem{survey_autonomous_driving}
J.~Mao, S.~Shi, X.~Wang, and H.~Li, ``{3D} object detection for autonomous driving: A comprehensive survey,'' \emph{Int. J. Comput. Vis.}, vol. 131, no.~8, pp. 1909--1963, 2023.

\bibitem{survey_4d_radar}
L.~Fan, J.~Wang, Y.~Chang, Y.~Li, Y.~Wang, and D.~Cao, ``{4D} mmwave radar for autonomous driving perception: A comprehensive survey,'' \emph{{IEEE} Trans. Intell. Veh.}, vol.~9, no.~4, pp. 4606--4620, 2024.

\bibitem{sgdet3d}
X.~Bai, Z.~Yu, L.~Zheng, X.~Zhang, Z.~Zhou, X.~Zhang, F.~Wang, J.~Bai, and H.-L. Shen, ``{SGDet3D}: Semantics and geometry fusion for {3D} object detection using {4D} radar and camera,'' \emph{{IEEE} Robot. Autom. Lett.}, 2024.

\bibitem{sfgfusion}
X.~Li, H.~Di, J.~Li, F.~Liu, and W.~Liang, ``{SFGFusion}: Surface fitting guided {3D} object detection with {4D} radar and camera fusion,'' \emph{arXiv preprint arXiv:2510.19215}, 2025.

\bibitem{hgsfusion}
Z.~Gu, J.~Ma, Y.~Huang, H.~Wei, Z.~Chen, H.~Zhang, and W.~Hong, ``{HGSFusion}: Radar-camera fusion with hybrid generation and synchronization for {3D} object detection,'' in \emph{Proc. AAAI Conf. Artif. Intell.}, vol.~39, no.~3, 2025, pp. 3185--3193.

\bibitem{survey_robust}
Z.~Song, L.~Liu, F.~Jia, Y.~Luo, C.~Jia, G.~Zhang, L.~Yang, and L.~Wang, ``Robustness-aware {3D} object detection in autonomous driving: A review and outlook,'' \emph{{IEEE} Trans. Intell. Transp. Syst.}, vol.~25, no.~11, pp. 15\,407--15\,436, 2024.

\bibitem{sequential_survey}
H.~Wang and Y.~Tian, ``Sequential point clouds: A survey,'' \emph{{IEEE} Trans. Pattern Anal. Mach. Intell.}, vol.~46, no.~8, pp. 5504--5523, 2024.

\bibitem{centerpoint}
T.~Yin, X.~Zhou, and P.~Krahenbuhl, ``Center-based {3D} object detection and tracking,'' in \emph{Proc. IEEE/CVF Conf. Comput. Vis. Pattern Recognit. (CVPR)}, 2021, pp. 11\,784--11\,793.

\bibitem{embracing}
L.~Fan, Z.~Pang, T.~Zhang, Y.-X. Wang, H.~Zhao, F.~Wang, N.~Wang, and Z.~Zhang, ``Embracing single stride 3d object detector with sparse transformer,'' in \emph{Proc. IEEE/CVF Conf. Comput. Vis. Pattern Recognit. (CVPR)}, 2022, pp. 8458--8468.

\bibitem{gmpnet-pami}
J.~Yin, J.~Shen, X.~Gao, D.~J. Crandall, and R.~Yang, ``Graph neural network and spatiotemporal transformer attention for 3d video object detection from point clouds,'' \emph{{IEEE} Trans. Pattern Anal. Mach. Intell.}, vol.~45, no.~8, pp. 9822--9835, 2021.

\bibitem{centerformer}
Z.~Zhou, X.~Zhao, Y.~Wang, P.~Wang, and H.~Foroosh, ``Centerformer: Center-based transformer for {3D} object detection,'' in \emph{Proc. Eur. Conf. Comput. Vis. (ECCV)}.\hskip 1em plus 0.5em minus 0.4em\relax Springer, 2022, pp. 496--513.

\bibitem{stemd}
Y.~Zhang, Z.~Zhu, J.~Hou, and D.~Wu, ``Spatial-temporal graph enhanced detr towards multi-frame {3D} object detection,'' \emph{{IEEE} Trans. Pattern Anal. Mach. Intell.}, vol.~46, no.~12, pp. 10\,614--10\,628, 2024.

\bibitem{mppnet}
X.~Chen, S.~Shi, B.~Zhu, K.~C. Cheung, H.~Xu, and H.~Li, ``{MPPNet}: Multi-frame feature intertwining with proxy points for {3D} temporal object detection,'' in \emph{Proc. Eur. Conf. Comput. Vis. (ECCV)}.\hskip 1em plus 0.5em minus 0.4em\relax Springer, 2022, pp. 680--697.

\bibitem{once-detected}
L.~Fan, Y.~Yang, Y.~Mao, F.~Wang, Y.~Chen, N.~Wang, and Z.~Zhang, ``Once detected, never lost: Surpassing human performance in offline lidar based {3D} object detection,'' in \emph{Proc. IEEE/CVF Conf. Comput. Vis. Pattern Recognit. (CVPR)}, 2023, pp. 19\,820--19\,829.

\bibitem{detzero}
T.~Ma, X.~Yang, H.~Zhou, X.~Li, B.~Shi, J.~Liu, Y.~Yang, Z.~Liu, L.~He, Y.~Qiao \emph{et~al.}, ``Detzero: Rethinking offboard {3D} object detection with long-term sequential point clouds,'' in \emph{Proc. IEEE/CVF Conf. Comput. Vis. Pattern Recognit. (CVPR)}, 2023, pp. 6736--6747.

\bibitem{fusionformer}
C.~Hu, H.~Zheng, K.~Li, J.~Xu, W.~Mao, M.~Luo, L.~Wang, M.~Chen, Q.~Peng, K.~Liu \emph{et~al.}, ``{FusionFormer}: A multi-sensory fusion in bird's-eye-view and temporal consistent transformer for {3D} object detection,'' \emph{arXiv preprint arXiv:2309.05257}, 2023.

\bibitem{bevfusion4d}
H.~Cai, Z.~Zhang, Z.~Zhou, Z.~Li, W.~Ding, and J.~Zhao, ``{Bevfusion4D}: Learning lidar-camera fusion under bird's-eye-view via cross-modality guidance and temporal aggregation,'' \emph{arXiv preprint arXiv:2303.17099}, 2023.

\bibitem{dmfusion}
X.~Yu, K.~Lu, Y.~Yang, and L.~Ou, ``{DMFusion}: {LiDAR-camera} fusion framework with depth merging and temporal aggregation,'' \emph{Appl. Intell.}, vol.~54, no.~19, pp. 9412--9428, 2024.

\bibitem{lstm}
S.~Hochreiter and J.~Schmidhuber, ``Long short-term memory,'' \emph{Neural Comput.}, vol.~9, no.~8, pp. 1735--1780, 1997.

\bibitem{attention}
A.~Vaswani, N.~Shazeer, N.~Parmar, J.~Uszkoreit, L.~Jones, A.~N. Gomez, {\L}.~Kaiser, and I.~Polosukhin, ``Attention is all you need,'' \emph{Adv. Neural Inf. Proces. Syst. (NeurIPS)}, vol.~30, 2017.

\bibitem{imvoxelnet}
D.~Rukhovich, A.~Vorontsova, and A.~Konushin, ``{ImVoxelNet}: Image to voxels projection for monocular and multi-view general-purpose {3D} object detection,'' in \emph{Proc. IEEE/CVF Winter Conf. Appl. Comput. Vis. Workshops (WACVW)}, 2022, pp. 2397--2406.

\bibitem{BEVRefiner}
B.~Wang, H.~Zheng, L.~Zhang, N.~Liu, R.~M. Anwer, H.~Cholakkal, Y.~Zhao, and Z.~Li, ``{BEVRefiner}: Improving {3D} object detection in bird’s-eye-view via dual refinement,'' \emph{{IEEE} Trans. Intell. Transp. Syst.}, vol.~25, no.~10, pp. 15\,094--15\,105, 2024.

\bibitem{pointpillars}
A.~H. Lang, S.~Vora, H.~Caesar, L.~Zhou, J.~Yang, and O.~Beijbom, ``{PointPillars}: Fast encoders for object detection from point clouds,'' in \emph{Proc. IEEE/CVF Conf. Comput. Vis. Pattern Recognit. (CVPR)}, 2019, pp. 12\,697--12\,705.

\bibitem{RPFA-Net}
B.~Xu, X.~Zhang, L.~Wang, X.~Hu, Z.~Li, S.~Pan, J.~Li, and Y.~Deng, ``{RPFA-Net}: A {4D} radar pillar feature attention network for {3D} object detection,'' in \emph{IEEE Conf. Intell. Transport. Syst. Proc. (ITSC)}, 2021, pp. 3061--3066.

\bibitem{smurf}
J.~Liu, Q.~Zhao, W.~Xiong, T.~Huang, Q.-L. Han, and B.~Zhu, ``{SMURF}: Spatial multi-representation fusion for {3D} object detection with {4D} imaging radar,'' \emph{{IEEE} Trans. Intell. Veh.}, vol.~9, no.~1, pp. 799--812, 2024.

\bibitem{pv_enconet}
Z.~Ouyang, X.~Dong, J.~Cui, J.~Niu, and M.~Guizani, ``{PV-EncoNet}: Fast object detection based on colored point cloud,'' \emph{{IEEE} Trans. Intell. Transp. Syst.}, vol.~23, no.~8, pp. 12\,439--12\,450, 2021.

\bibitem{virtual-sparse}
H.~Wu, C.~Wen, S.~Shi, X.~Li, and C.~Wang, ``Virtual sparse convolution for multimodal {3D} object detection,'' in \emph{Proc. IEEE/CVF Conf. Comput. Vis. Pattern Recognit. (CVPR)}, 2023, pp. 21\,653--21\,662.

\bibitem{RCBevdet}
Z.~Lin, Z.~Liu, Z.~Xia, X.~Wang, Y.~Wang, S.~Qi, Y.~Dong, N.~Dong, L.~Zhang, and C.~Zhu, ``{RCBEVDet}: Radar-camera fusion in bird's eye view for {3D} object detection,'' in \emph{Proc. IEEE/CVF Conf. Comput. Vis. Pattern Recognit. (CVPR)}, 2024, pp. 14\,928--14\,937.

\bibitem{pointaugmenting}
C.~Wang, C.~Ma, M.~Zhu, and X.~Yang, ``{PointAugmenting}: Cross-modal augmentation for {3D} object detection,'' in \emph{Proc. IEEE/CVF Conf. Comput. Vis. Pattern Recognit. (CVPR)}, 2021, pp. 11\,794--11\,803.

\bibitem{mv3d}
X.~Chen, H.~Ma, J.~Wan, B.~Li, and T.~Xia, ``Multi-view {3D} object detection network for autonomous driving,'' in \emph{Proc. IEEE/CVF Conf. Comput. Vis. Pattern Recognit. (CVPR)}, 2017, pp. 1907--1915.

\bibitem{f-pointnet}
C.~R. Qi, W.~Liu, C.~Wu, H.~Su, and L.~J. Guibas, ``Frustum pointnets for {3D} object detection from {RGB-D} data,'' in \emph{Proc. IEEE/CVF Conf. Comput. Vis. Pattern Recognit. (CVPR)}, 2018, pp. 918--927.

\bibitem{cl3d}
C.~Lin, D.~Tian, X.~Duan, J.~Zhou, D.~Zhao, and D.~Cao, ``{CL3D}: Camera-lidar {3D} object detection with point feature enhancement and point-guided fusion,'' \emph{{IEEE} Trans. Intell. Transp. Syst.}, vol.~23, no.~10, pp. 18\,040--18\,050, 2022.

\bibitem{bevfusion_NeurIPS}
T.~Liang, H.~Xie, K.~Yu, Z.~Xia, Z.~Lin, Y.~Wang, T.~Tang, B.~Wang, and Z.~Tang, ``Bevfusion: A simple and robust lidar-camera fusion framework,'' \emph{Adv. Neural Inf. Proces. Syst. (NeurIPS)}, vol.~35, pp. 10\,421--10\,434, 2022.

\bibitem{bevfusion}
Z.~Liu, H.~Tang, A.~Amini, X.~Yang, H.~Mao, D.~L. Rus, and S.~Han, ``{BEVFusion}: Multi-task multi-sensor fusion with unified bird's-eye view representation,'' in \emph{Proc. IEEE Int. Conf. Rob. Autom. (ICRA)}, 2023, pp. 2774--2781.

\bibitem{lss}
J.~Philion and S.~Fidler, ``Lift, splat, shoot: Encoding images from arbitrary camera rigs by implicitly unprojecting to {3D},'' in \emph{Proc. Eur. Conf. Comput. Vis. (ECCV)}, 2020, pp. 194--210.

\bibitem{3d-man}
Z.~Yang, Y.~Zhou, Z.~Chen, and J.~Ngiam, ``{3D-MAN}: {3D} multi-frame attention network for object detection,'' in \emph{Proc. IEEE/CVF Conf. Comput. Vis. Pattern Recognit. (CVPR)}, 2021, pp. 1863--1872.

\bibitem{4d-net}
A.~Piergiovanni, V.~Casser, M.~S. Ryoo, and A.~Angelova, ``{4D-Net} for learned multi-modal alignment,'' in \emph{Proc. IEEE/CVF Int. Conf. Comput. Vis. (ICCV)}, 2021, pp. 15\,435--15\,445.

\bibitem{deepfusion}
Y.~Li, A.~W. Yu, T.~Meng, B.~Caine, J.~Ngiam, D.~Peng, J.~Shen, Y.~Lu, D.~Zhou, Q.~V. Le \emph{et~al.}, ``Deepfusion: Lidar-camera deep fusion for multi-modal 3d object detection,'' in \emph{Proc. IEEE/CVF Conf. Comput. Vis. Pattern Recognit. (CVPR)}, 2022, pp. 17\,182--17\,191.

\bibitem{lift}
Y.~Zeng, D.~Zhang, C.~Wang, Z.~Miao, T.~Liu, X.~Zhan, D.~Hao, and C.~Ma, ``Lift: Learning 4d lidar image fusion transformer for 3d object detection,'' in \emph{CVPR}, 2022, pp. 17\,172--17\,181.

\bibitem{immortal}
Q.~Wang, Y.~Chen, Z.~Pang, N.~Wang, and Z.~Zhang, ``{Immortal Tracker}: Tracklet never dies,'' \emph{arXiv preprint arXiv:2111.13672}, 2021.

\bibitem{vod}
A.~Palffy, E.~Pool, S.~Baratam, J.~F. Kooij, and D.~M. Gavrila, ``Multi-class road user detection with 3+1{D} radar in the view-of-delft dataset,'' \emph{{IEEE} Robot. Autom. Lett.}, vol.~7, no.~2, pp. 4961--4968, 2022.

\bibitem{tj4dradset}
L.~Zheng, Z.~Ma, X.~Zhu, B.~Tan, S.~Li, K.~Long, W.~Sun, S.~Chen, L.~Zhang, M.~Wan \emph{et~al.}, ``{TJ4DRadSet}: A {4D} radar dataset for autonomous driving,'' in \emph{IEEE Conf. Intell. Transport. Syst. Proc. (ITSC)}, 2022, pp. 493--498.

\bibitem{pillarnext}
J.~Li, C.~Luo, and X.~Yang, ``{PillarNeXt}: Rethinking network designs for {3D} object detection in lidar point clouds,'' in \emph{Proc. IEEE/CVF Conf. Comput. Vis. Pattern Recognit. (CVPR)}, 2023, pp. 17\,567--17\,576.

\bibitem{RCFusion}
L.~Zheng, S.~Li, B.~Tan, L.~Yang, S.~Chen, L.~Huang, J.~Bai, X.~Zhu, and Z.~Ma, ``{RCFusion}: Fusing 4-{D} radar and camera with bird’s-eye view features for 3-{D} object detection,'' \emph{{IEEE} Trans. Instrum. Meas.}, vol.~72, pp. 1--14, 2023.

\bibitem{smiformer}
W.~Shi, Z.~Zhu, K.~Zhang, H.~Chen, Z.~Yu, and Y.~Zhu, ``{SMIFormer}: Learning spatial feature representation for {3D} object detection from {4D} imaging radar via multi-view interactive transformers,'' \emph{Sensors}, vol.~23, no.~23, p. 9429, 2023.

\bibitem{sckd}
R.~Xu, Z.~Xiang, C.~Zhang, H.~Zhong, X.~Zhao, R.~Dang, P.~Xu, T.~Pu, and E.~Liu, ``{SCKD}: Semi-supervised cross-modality knowledge distillation for {4D} radar object detection,'' in \emph{Proc. AAAI Conf. Artif. Intell.}, vol.~39, no.~9, 2025, pp. 8933--8941.

\bibitem{futr3d}
X.~Chen, T.~Zhang, Y.~Wang, Y.~Wang, and H.~Zhao, ``{FUTR3D}: A unified sensor fusion framework for {3D} detection,'' in \emph{Proc. IEEE/CVF Conf. Comput. Vis. Pattern Recognit. (CVPR)}, 2023, pp. 172--181.

\bibitem{LXL}
W.~Xiong, J.~Liu, T.~Huang, Q.-L. Han, Y.~Xia, and B.~Zhu, ``{LXL}: Lidar excluded lean {3D} object detection with {4D} imaging radar and camera fusion,'' \emph{{IEEE} Trans. Intell. Veh.}, pp. 3142--3142, 2024.

\bibitem{zfusion}
S.~Yang, T.~Zhan, S.~Qiao, J.~Gong, Q.~Yang, J.~Wang, and Y.~Lu, ``Zfusion: An effective fuser of camera and {4D} radar for {3D} object perception in autonomous driving,'' in \emph{Proc. IEEE/CVF Conf. Comput. Vis. Pattern Recognit. Workshops (CVPRW)}, June 2025, pp. 3768--3777.

\bibitem{lxlv2}
W.~Xiong, Z.~Zou, Q.~Zhao, F.~He, and B.~Zhu, ``{LXLv2}: Enhanced lidar excluded lean {3D} object detection with fusion of {4D} radar and camera,'' \emph{{IEEE} Robot. Autom. Lett.}, 2025.

\bibitem{doracamom}
L.~Zheng, J.~Liu, R.~Guan, L.~Yang, S.~Lu, Y.~Li, X.~Bai, J.~Bai, Z.~Ma, H.-L. Shen \emph{et~al.}, ``Doracamom: Joint {3D} detection and occupancy prediction with multi-view {4D} radars and cameras for omnidirectional perception,'' \emph{arXiv preprint arXiv:2501.15394}, 2025.

\bibitem{hydra}
P.~Wolters, J.~Gilg, T.~Teepe, F.~Herzog, A.~Laouichi, M.~Hofmann, and G.~Rigoll, ``Unleashing hydra: Hybrid fusion, depth consistency and radar for unified {3D} perception,'' in \emph{Proc. IEEE Int. Conf. Rob. Autom. (ICRA)}.\hskip 1em plus 0.5em minus 0.4em\relax IEEE, 2025, pp. 7467--7474.

\bibitem{mssf}
H.~Liu, J.~Liu, G.~Jiang, and X.~Jin, ``Mssf: A {4D} radar and camera fusion framework with multi-stage sampling for {3D} object detection in autonomous driving,'' \emph{{IEEE} Trans. Intell. Transp. Syst.}, 2025.

\bibitem{openpcdet2020}
O.~D. Team, ``{OpenPCDet}: An open-source toolbox for {3D} object detection from point clouds,'' \url{https://github.com/open-mmlab/OpenPCDet}, 2020.

\bibitem{voxel_mamba}
G.~Zhang, L.~Fan, C.~He, Z.~Lei, Z.~Zhang, and L.~Zhang, ``Voxel mamba: Group-free state space models for point cloud based {3D} object detection,'' \emph{Adv. Neural Inf. Proces. Syst. (NeurIPS)}, vol.~37, pp. 81\,489--81\,509, 2024.

\bibitem{mvx-net}
V.~A. Sindagi, Y.~Zhou, and O.~Tuzel, ``{MVX-Net}: Multimodal voxelnet for {3D} object detection,'' in \emph{Proc. IEEE Int. Conf. Rob. Autom. (ICRA)}.\hskip 1em plus 0.5em minus 0.4em\relax IEEE, 2019, pp. 7276--7282.

\bibitem{interfusion}
L.~Wang, X.~Zhang, B.~Xv, J.~Zhang, R.~Fu, X.~Wang, L.~Zhu, H.~Ren, P.~Lu, J.~Li \emph{et~al.}, ``{InterFusion}: Interaction-based {4D} radar and lidar fusion for {3D} object detection,'' in \emph{IEEE Int. Conf. Intell. Rob. Syst. (IROS)}.\hskip 1em plus 0.5em minus 0.4em\relax IEEE, 2022, pp. 12\,247--12\,253.

\bibitem{second}
Y.~Yan, Y.~Mao, and B.~Li, ``{SECOND}: Sparsely embedded convolutional detection,'' \emph{Sensors}, vol.~18, no.~10, p. 3337, 2018.

\bibitem{part-A}
S.~Shi, Z.~Wang, J.~Shi, X.~Wang, and H.~Li, ``From points to parts: {3D} object detection from point cloud with part-aware and part-aggregation network,'' \emph{{IEEE} Trans. Pattern Anal. Mach. Intell.}, vol.~43, no.~8, pp. 2647--2664, 2020.

\bibitem{focal}
Y.~Chen, Y.~Li, X.~Zhang, J.~Sun, and J.~Jia, ``Focal sparse convolutional networks for 3d object detection,'' in \emph{Proc. IEEE/CVF Conf. Comput. Vis. Pattern Recognit. (CVPR)}, 2022, pp. 5428--5437.

\end{thebibliography}
\vspace{-10 mm}
\begin{IEEEbiography}[{\includegraphics[width=1in,height=1.25in,clip,keepaspectratio]{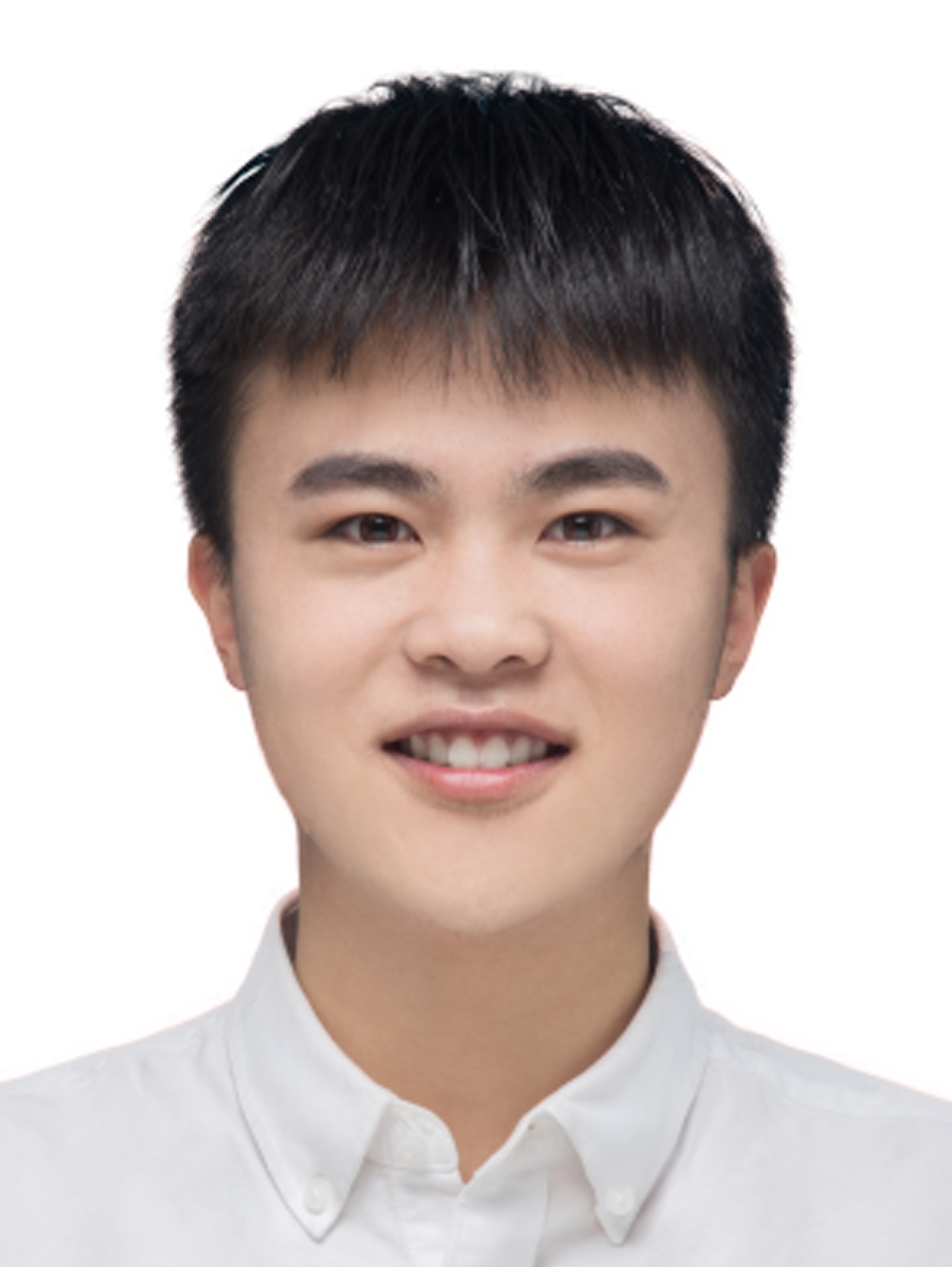}}]{Xiaozhi Li} 
received the B.S. degree from the Hefei University of Technology, Hefei, China, in 2020, and the M.S. degree from the Jilin University, Changchun, China, in 2023. He is currently pursuing the Ph.D. degree with the School of Information and Electronics, Beijing Institute of Technology, Beijing, China. His research interests include deep learning, computer vision, object detection, and multi-modal environment perception.
\end{IEEEbiography}
\vspace{-10 mm}
\begin{IEEEbiography}
[{\includegraphics[width=1in,height=1.25in,clip,keepaspectratio]{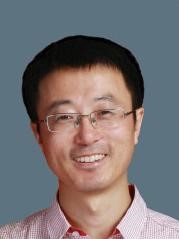}}]{Huijun Di} 
is an Associate Professor at the School of Computer Science at Beijing Institute of Technology, China. He received the B.S. degree in 2002 and the Ph.D. degree in 2009, both from Tsinghua University, China. He was a postdoctoral researcher in the Department of Computer Science and Technology at Tsinghua University from 2009 to 2012.  His main research interests include computer vision, artificial intelligence, pattern recognition, machine learning, and intelligent systems.
\end{IEEEbiography}
\vspace{-10 mm}
\begin{IEEEbiography}
[{\includegraphics[width=1in,height=1.25in,clip,keepaspectratio]{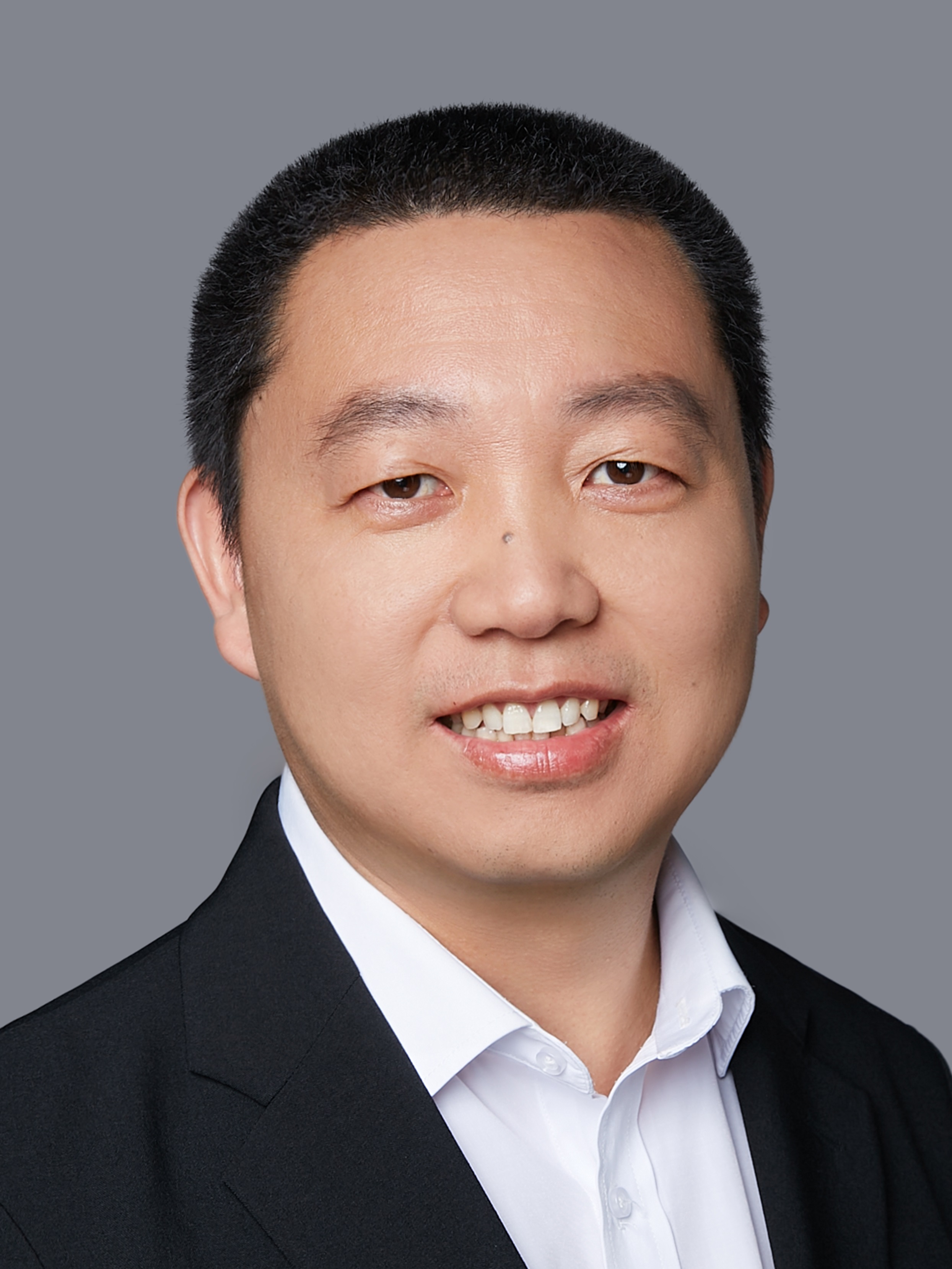}}]{Jian Li} 
received the B.S. degree in 2004 and the Ph.D. degree in 2009, both from Beijing Institute of Technology, Beijing, China. He is currently a Lecturer at the Key Laboratory of Electronic Information Technology for Satellite Navigation, Ministry of Education, Beijing Institute of Technology. He has long been dedicated to research in the fields of satellite navigation signal processing and multi-source information fusion. 
\end{IEEEbiography}
\vspace{-10 mm}
\begin{IEEEbiography}
[{\includegraphics[width=1in,height=1.25in,clip,keepaspectratio]{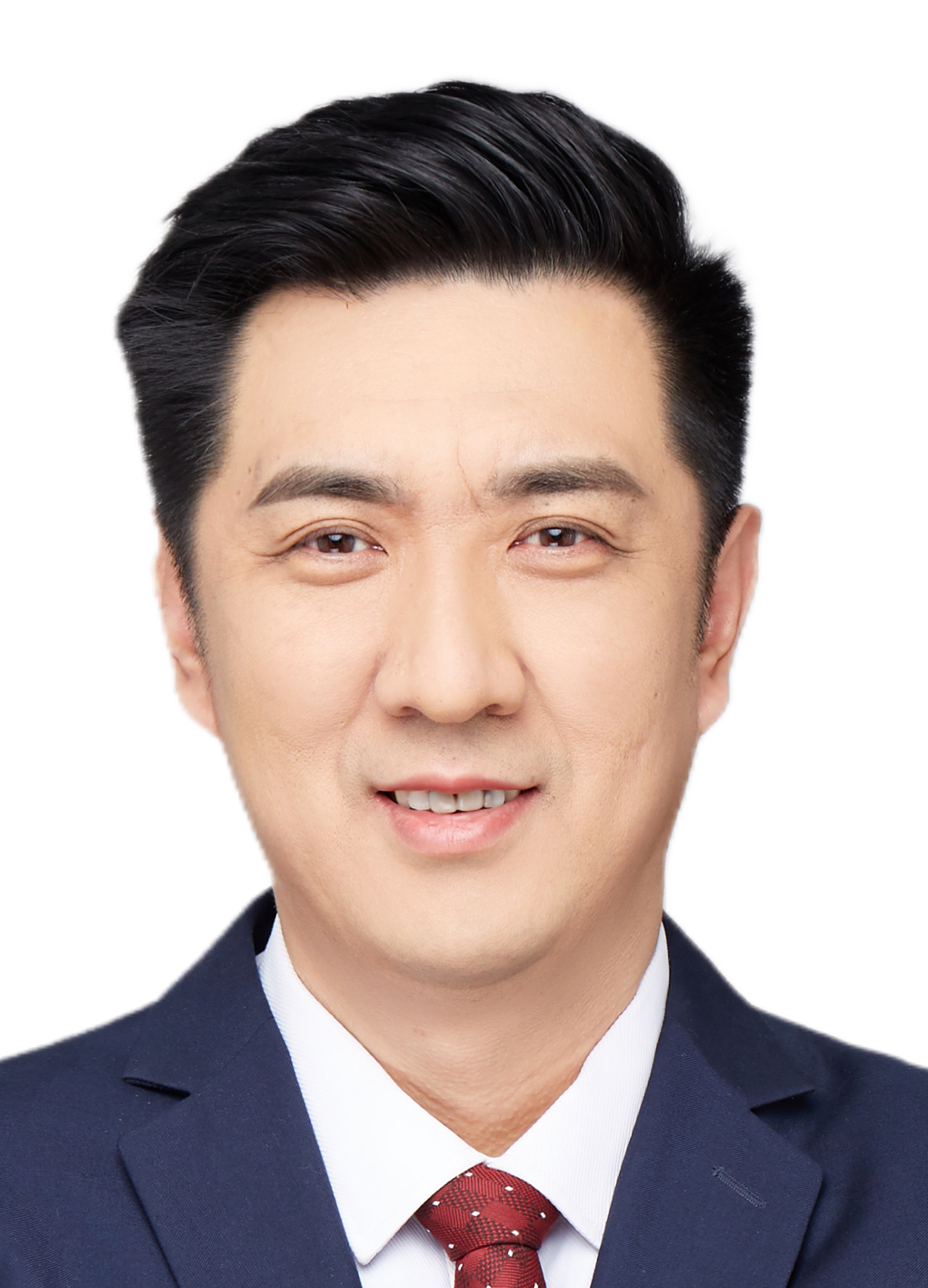}}]{Feng Liu} 
received the B.S. degree in 1999 and the Ph.D. degree in 2004, both from Beijing Institute of Technology, Beijing, China. He is currently the General Manager of Beijing Racobit Electronic Information Technology Co., Ltd. His research focuses on satellite navigation receivers, various radar systems, and intelligent transportation systems.
\end{IEEEbiography}
\vspace{-10 mm}
\begin{IEEEbiography}
[{\includegraphics[width=1in,height=1.25in,clip,keepaspectratio]{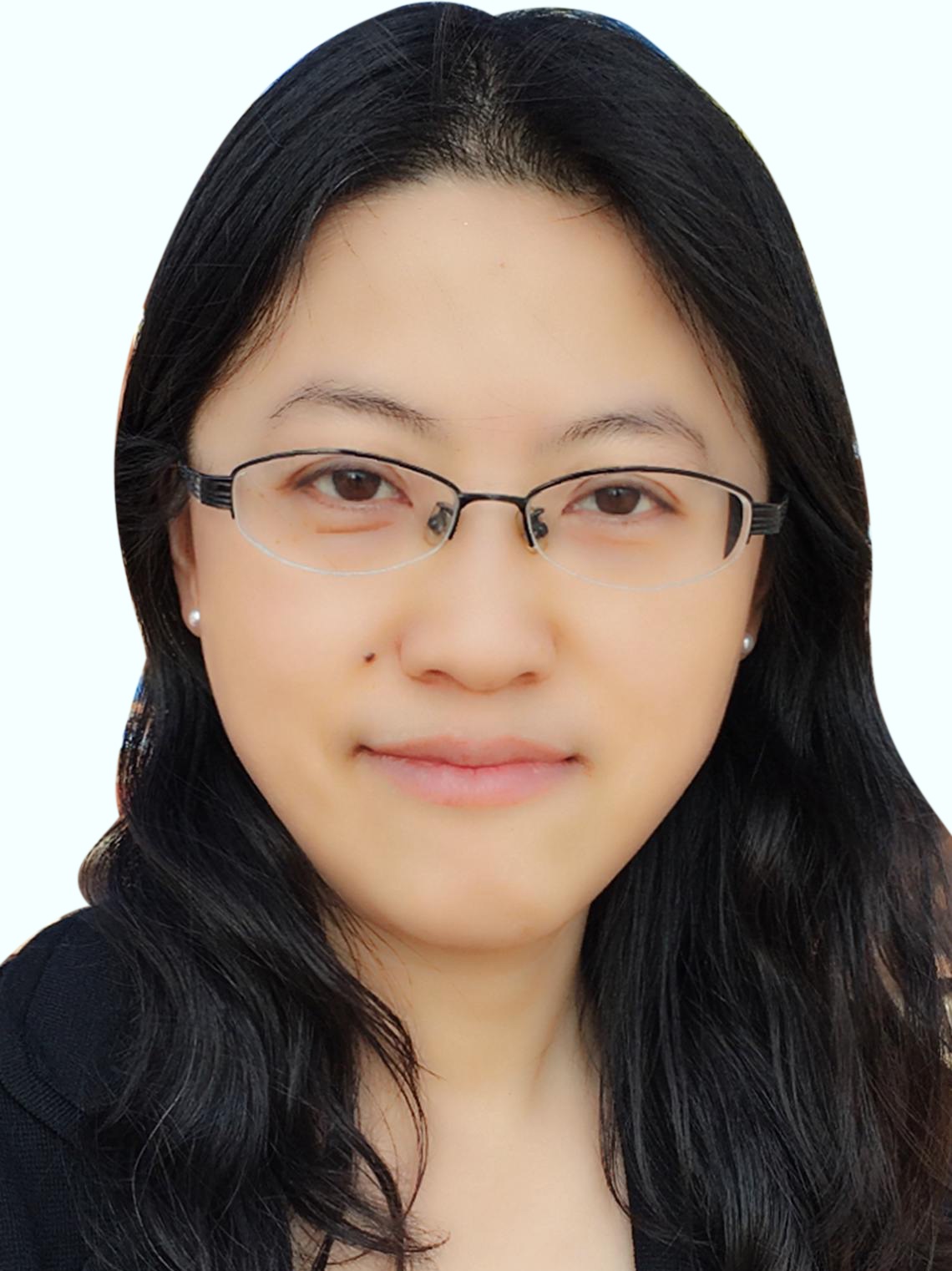}}]{Wei Liang} 
is a professor at the School of Computer Science and Technology at Beijing Institute of Technology (BIT), China. She earned her Ph.D. in Computer Science from BIT in 2005. In 2014, she was a visiting professor at the Vision Cognition Learning and Autonomy (VCLA) Lab at UCLA, USA. Her research interests focus on Computer Vision and Human-Computer Interaction (HCI).
\end{IEEEbiography}

\end{document}